\documentclass[10pt,twocolumn,letterpaper]{article}

\usepackage[pagenumbers]{cvpr}

\usepackage[accsupp]{axessibility}
\usepackage{graphicx}
\usepackage{amsmath}
\usepackage{amssymb}
\usepackage{booktabs}
\usepackage{pifont}
\usepackage{color}
\usepackage{xcolor}
\usepackage{bm}
\usepackage{multirow}

\usepackage[moderate]{savetrees}

\definecolor{citecolor}{HTML}{0071bc}
\usepackage[pagebackref=true,breaklinks=true,colorlinks,citecolor=citecolor,bookmarks=false]{hyperref}

\usepackage[capitalize]{cleveref}
\crefname{section}{Sec.}{Secs.}
\Crefname{section}{Section}{Sections}
\Crefname{table}{Table}{Tables}
\crefname{table}{Tab.}{Tabs.}

\def\cvprPaperID{****}
\def\confName{****}
\def\confYear{****}

\definecolor{green}{RGB}{4,255,40}
\definecolor{red}{RGB}{255,0,0}
\definecolor{tgreen}{RGB}{4,155,70}
\definecolor{purple}{rgb}{0.896, 0.39, 1.0}

\newlength\savewidth\newcommand\shline{\noalign{\global\savewidth\arrayrulewidth
  \global\arrayrulewidth 1pt}\hline\noalign{\global\arrayrulewidth\savewidth}}
\newcommand{\tablestyle}[2]{\setlength{\tabcolsep}{#1}\renewcommand{\arraystretch}{#2}\centering\footnotesize}
\newcommand{\cmark}{\ding{51}}
\newcommand{\xmark}{\ding{55}}

\begin{document}

\title{Joint Hand Motion and Interaction Hotspots Prediction from Egocentric Videos}

\author{Shaowei Liu\textsuperscript{1}\thanks{Work partially done during an internship at Intel Labs.} \quad\quad 
Subarna Tripathi\textsuperscript{2} \quad\quad 
Somdeb Majumdar\textsuperscript{2} \quad\quad
Xiaolong Wang\textsuperscript{3}\\
\textsuperscript{1}University of Illinois Urbana-Champaign \qquad \textsuperscript{2} Intel Labs \qquad \textsuperscript{3}UC San Diego
}
\maketitle


\begin{abstract}
We propose to forecast future hand-object interactions given an egocentric video. Instead of predicting action labels or pixels, we directly predict the hand motion trajectory and the future contact points on the next active object (i.e., interaction hotspots). This relatively low-dimensional representation provides a concrete description of future interactions. To tackle this task, we first provide an automatic way to collect trajectory and hotspots labels on large-scale data. We then use this data to train an Object-Centric Transformer (OCT) model for prediction. Our model performs hand and object interaction reasoning via the self-attention mechanism in Transformers. OCT also provides a probabilistic framework to sample the future trajectory and hotspots to handle uncertainty in prediction. We perform experiments on the Epic-Kitchens-55, Epic-Kitchens-100 and EGTEA Gaze+ datasets, and show that OCT significantly outperforms state-of-the-art approaches by a large margin. 
Project page is available at \href{https://stevenlsw.github.io/hoi-forecast}{https://stevenlsw.github.io/hoi-forecast}.
\end{abstract}

\section{Introduction}
\label{sec:intro}

\begin{figure}[t]
\centering
\includegraphics[width=\linewidth]{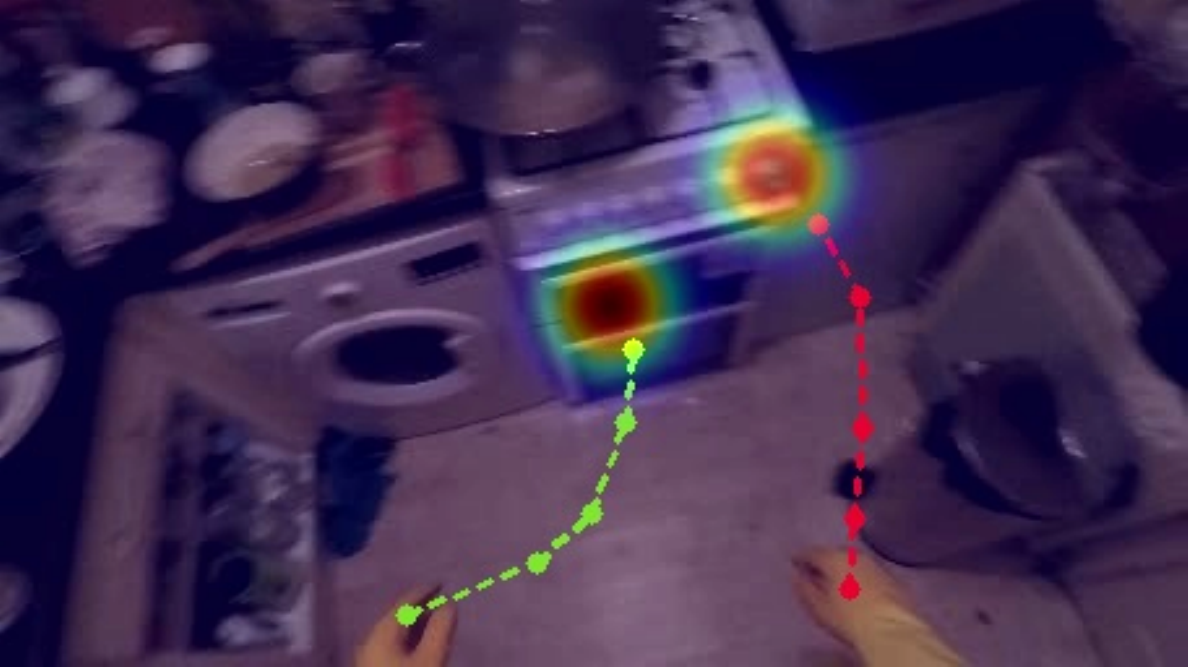}
\caption{Going beyond predicting a single action label in the future, we propose to jointly predict the future hand motion trajectories (blue and red lines) and interaction hotspots (heatmaps) on the next-active object in egocentric videos.}
\vspace{-0.2in}
\label{fig:teaser}
\end{figure}

Achieving the ability to predict a person's intent, preference and future activities is one of the fundamental goals for AI systems. This is particularly useful when it comes to egocentric video data for applications such as augmented reality (AR) and robotics. Imagining with an egocentric view inside the kitchen (e.g., Figure~\ref{fig:teaser}), if an AI system can forecast what the human would do next, an AR headset could provide useful and timely guidance, and a robot can react and collaborate with the human more smoothly. 

What space should the model predict on? Recent approaches~\cite{furnari2019would, furnari2020rolling, sener2020temporal, girdhar2021anticipative} have been proposed to predict the discrete future action category given a sequence of frames as inputs, namely action anticipation. However, predicting a semantic label does not reveal how the human moves and what the human will interact with in the future. On the other hand, predicting pixels for future frames~\cite{lotter2016deep, byeon2018contextvp, wang2018eidetic, lee2021video} is very challenging due to its high dimension outputs with large uncertainties. Instead of adopting these two representations, our work is inspired by recent work on human motion trajectory prediction~\cite{cao2020long} which takes images as inputs and outputs the coordinates of future pose joints. Trajectory not only provides a concrete description of motion, but also is a much smaller space to predict compared to pixel prediction. However, unlike previous works, prediction in egocentric videos also involves dense interactions with objects, which cannot be modeled by trajectory alone. 

In this paper, we propose to jointly predict the future hand motion trajectory and the interaction hotspots (affordance) of the next-active object, given a sequence of input frames from an egocentric video. Starting from the final frame of the input video, we will predict the trajectories for both hands by sampling from a probabilistic distribution inferred by the model. Instead of learning a deterministic model, we tackle the uncertainty of future in a probabilistic manner. At the same time, we will predict the contact points on the next-active object interacted by the future hands. These contact points are also represented via probabilistic distributions in the form of interaction hotspots~\cite{nagarajan2019grounded} and conditioned on the predicted hand trajectory. To perform joint predictions, we introduce a Transformer-based model and an automatic way to generate a large-scale dataset for training. 

Instead of collecting annotations for hand trajectories and interaction hotspots with high-cost human labor, we propose an automatic manner to collect the data in a large-scale. Given a video, we call the input frames to our model the observation frames and the predicted ones are called future frames. We first utilize off-the-shelf hand detectors~\cite{shan2020understanding} to locate hands in all the future frames. Since the camera is usually moving in egocentric videos, we leverage homography in nearby frames and project the detected future hands' locations
back to the last observation frame. In this way, all the detections are aligned in the same coordinate system. Similarly, we also detect the locations where the hand interacts with the object in future frames, and project them back to the last observation frame. This process prepares the data for training our prediction model, and we generate labels for  Epic-Kitchens-55, Epic-Kitchens-100 and EGTEA Gaze+ datasets without any human labor.

With the collected data, we propose to learn a novel Object-Centric Transformer (OCT) model which captures the hand-object relations from videos for hand trajectory and interaction hotspots prediction. Given the observation frames as inputs, we first extract their visual representations with a ConvNet. We perform hand and object detection and adopt RoI Align~\cite{he2017mask} to extract their features. We take both hand and object features as object-centric tokens, and the average-pooled frame feature as image context tokens. We forward all tokens from all input frames to a Transformer encoder, which performs hand, object and environment context interaction reasoning using self-attention. Instead of decoding in a deterministic manner, we adopt the Conditional Variational Autoencoders (C-VAE) as network head in the Transformer decoder to model the uncertainty in prediction. Specifically, we compute cross-attention between the output tokens from the Transformer encoder and predicted future hand locations in the Transformer decoder. The obtained tokens are taken as conditional variables for the C-VAE. The decoder will predict both the hand trajectories and interaction hotspots jointly, and the training is supervised by a reconstruction loss corresponding to the ground-truths.

We perform evaluation on Epic-Kitchens-55~\cite{damen2018scaling}, Epic-Kitchens-100~\cite{damen2021rescaling} and EGTEA Gaze+~\cite{li2018eye} datasets. We manually annotate the validation sets with trajectory and hotspots labels using the Amazon Mechanical Turk platform. Our OCT model significantly outperforms the baselines on both hand trajectory and interaction hotspots prediction tasks. Interestingly, we find that trajectory estimation helps interaction hotspots prediction and with more automatic annotated training data we can get better results. Finally, we experiment with fine-tuning the trained model on the action anticipation task, and find that predicting hand trajectory and interaction hotspots can benefit classifying future actions.

Our contributions are the following:
\begin{itemize}
    \item We propose to jointly predict hand trajectory and interaction hotspots from egocentric videos, and collect new training and test annotations.
    \item A novel Object-Centric Transformer which models the hand and object interactions for predicting future trajectory and affordance.
    \item We not only achieve state-of-the-art performance on both prediction tasks on Epic-Kitchens and EGTEA Gaze+ datasets, but also show our model can help the action anticipation task. 
\end{itemize}

\begin{figure*}[t]
    \centering
    \includegraphics[width=0.95\linewidth]{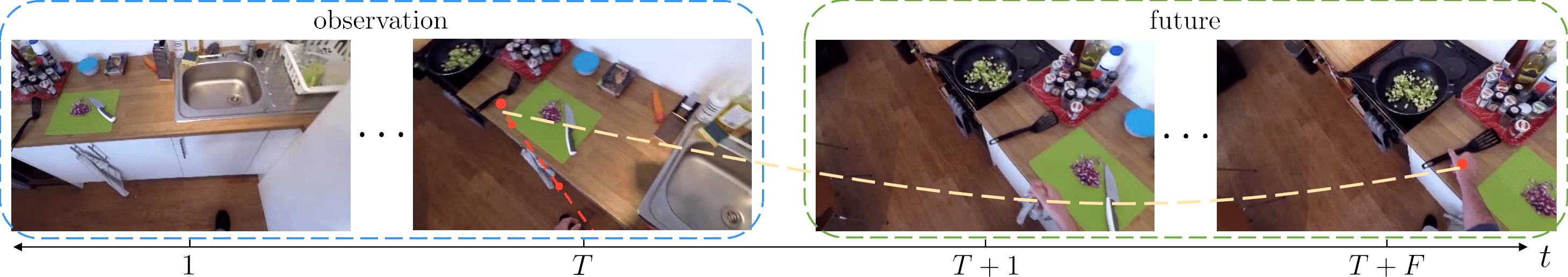}
    \vspace{-0.1in}
    \caption{Given $T$ observation frames as input (left), the goal is to forecast $F$ steps future hand trajectory (right) and interaction hotspots. The orange curve shows how we project future hand locations (red dots) to the last observation frame. The future hand trajectory is shown in red dashed line.}
    \label{fig: task}
\end{figure*}

\section{Related Work}
\textbf{Video anticipation.} Video anticipation aims to forecast future events
in videos, including future frames prediction~\cite{lotter2016deep, vondrick2016anticipating, byeon2018contextvp, wang2018eidetic, jayaraman2018time, ye2019compositional, lee2021video}, action anticipation~\cite{furnari2019would, miech2019leveraging, wu2020learning, furnari2020rolling, sener2020temporal, girdhar2021anticipative}, and dynamics learning~\cite{finn2016unsupervised, girdhar2020forward, qi2020learning}. However, most of these works either relied on anticipating high-dimensional visual representations of the future, which is extremely challenging in dynamic scenes with appearance changes and moving agents, or focused on predicting a semantic label of future actions. The labels could neither tell us where the person intends to move nor the object the person would like to interact with. On the contrary, we predict the future hand motion trajectory and the interaction hotspots, both reflecting human intention and future interactions in low dimensions.

\textbf{Human motion forecasting.} Predicting future human motions~\cite{li2017auto, habibie2017recurrent, cao2020long, petrovich2021action, wang2021synthesizing} or trajectories~\cite{alahi2016social, ma2017forecasting, liang2019peeking, mangalam2020not, dendorfer2020goal, mangalam2021goals, liu2020forecasting} has been a long-standing research topic. Many of them operate on third-person vision or fixed bird’s eye view settings. Given that the first-person vision could better capture people’s intention and interactions, as well as its applicability to AR and robotics~\cite{kanade2012first, koppula2015anticipating, rhinehart2018first}, estimating human motions in egocentric videos~\cite{park2016egocentric, yagi2018future, liu2020forecasting} worth more attention. As hands are central means for humans to explore and manipulate in egocentric videos, forecasting where human hands move could reveal future activity and understand a person's intention. Liu \textit{et al.}~\cite{liu2020forecasting} also studied future hand trajectory estimation in egocentric videos, but their method is limited by manual annotation and single hand prediction. In contrast, we design an automatic way to collect the data on a large scale and can learn future trajectories for both hands from the data.

\textbf{Grounded affordance prediction.} Object affordance grounding~\cite{kjellstrom2011visual, song2013predicting, myers2015affordance, sawatzky2017weakly, do2018affordancenet, Fang_2018_CVPR} refers to locating where the interaction occurs on an object. Given the video input, the affordance prediction task is to estimate the future active regions on an object that the human would interact. In general, there are two main categories of prediction, next-active object~\cite{bertasius2016first, furnari2017next, dessalene2021forecasting} and interaction hotspots~\cite{nagarajan2019grounded, liu2020forecasting, luo2021learning}. The former one segments the object that will next come into contact with the hand holistically, neglecting the fine-grained spatial regions on the object's surface. The latter one outputs a heatmap to indicate salient regions on the object. Nagarajan~\textit{et al.}~\cite{nagarajan2019grounded} proposed a weakly-supervised method to ground interaction hotspots on inactive images. Going one step further, we consider predicting interaction hotspots in egocentric videos. The task is more challenging since it needs to localize the next-active-object in a cluttered scene before grounding the hotspots. In our work, instead of using heatmap representation, we directly predict the contact locations more compactly.

\textbf{Transformers for video forecasting.} Following the immense success of Transformer~\cite{vaswani2017attention} in natural language processing, recent studies showed its effectiveness in solving vision tasks~\cite{carion2020end, chen2020generative, dosovitskiy2020image, liu2021swin, touvron2021training}. 
The long-range reasoning and sequence modeling capability make Transformers suitable for video understanding~\cite{girdhar2019video, bertasius2021space, liu2021video, wu2021towards}. The TimeSformer~\cite{bertasius2021space} viewed the video as a sequence of patches and adopted divided space-time attention to capture spatial-temporal relations in videos. Transformers have also been widely used in video forecasting problems like action anticipation~\cite{girdhar2021anticipative}, trajectory estimation~\cite{yu2020spatio, giuliari2021transformer} and human motion prediction~\cite{petrovich2021action, li2021ai}. Recent works show promising results by incorporating VAE~\cite{kingma2013auto} into Transformers for generative modeling~\cite{jiang2020transformer, fang2021transformer, petrovich2021action}. In our work, we propose an Object-Centric Transformer (OCT) that takes the RoIAlign~\cite{he2017mask} hand, object, and environment feature vectors extracted from a pre-trained ConvNet~\cite{wang2016temporal} as input tokens. The Transformer encoder adopts self-attention across all input tokens while the Transformer decoder computes cross-attention between output tokens from the encoder and predicted future hand locations. We also introduce the C-VAE~\cite{kingma2013auto} head in the Transformer decoder to express the uncertainty of the future.

\section{Problem Setup}

\subsection{Task Description}
\label{sec: task}
Given observation key frames $\mathcal{V} = \{f_1, \cdots, f_T\}$ of length $T$ as input, where $f_T$ is the last observation frame, our goal is to predict future hand trajectories $\mathcal{H}$ and object contact points $\mathcal{O}$ in the future time horizons of $F$, as shown in Figure~\ref{fig: task}. $\mathcal{H} = \{h_{T+1}, \cdots, h_{T+F}\}$
denotes the future hand trajectories. 
At each time step $t$, the future hand location $h_t=(h_l^t, h_r^t)$ consists of the left and right hand 2D pixel locations in the last observation frame. Time step $T+F$ is when the hand-object contact occurs and frame $f_{T+F}$ is the contact frame. $\mathcal{O}=\{o_1, \cdots, o_N\}$ denotes the future object contact points, where $N$ is the maximum number of contact predictions and each element defines the 2D future contact location in the last observation frame.

\subsection{Training Data Generation}
\label{sec: generation}
We describe how to collect training labels of future hand trajectory $\mathcal{H}$ and object hotspots $\mathcal{O}$ from future key frames $\{f_{t+1}, \cdots, f_T\}$ automatically without manual labor. We first run an off-the-shelf active hand-object detector~\cite{shan2020understanding} to get hand and object bounding boxes per frame, providing the future hand locations in each frame. Then we project them back to the last observation frame to collect a complete future hand trajectory. See Figure~\ref{fig: task} for illustration. As shown in~\cite{wang2013action, nagarajan2020ego}, the global motion between two consecutive frames is usually small, and they can be related by a homography~\cite{szeliski2006image}. Given the homography between every two consecutive frames, we could build a chain and establish the relations of each future frame w.r.t. the last observation frame and project the future hand location back. In order to estimate the homography, we first exclude the moving objects, in particular the detected hands and objects in each frame. We mask out the corresponding location and find the correspondences between two frames outside the masked regions using SURF descriptor~\cite{bay2006surf}. We calculate the homography by sampling 4 points and applying RANSAC to maximize the number of inliers. 

Similarly, for collecting the hotspots labels, we perform an additional skin segmentation~\cite{saxen2014color} and fingertip detection~\footnote{https://www.computervision.zone/courses/finger-counter/} within the active intersection region of hand and object bounding boxes to obtain contact points. Then we adopt a similar technique as above to project the sampled contact points from the contact frame to the last observation frame. More detailed discussions are provided in the supplementary.

\subsection{Pre-processing}
\label{sec: preprocess}
Given the video clip of observation frames $\mathcal{V} = \{f_1, \cdots, f_T\}$, we extract per-frame features, $\{X_1, \cdots, X_T\}$. Each frame consists of three types of input feature tokens $X_t=(X_h^t, X_o^t, X_g^t)$, where$X_i^t$ represents the feature of $i$-th type in frame $t$. Subscripts $h,o,g$ refer to the hand, object, and global feature (environment context) vectors respectively. To this end, we first encode each frame $f_t$ using a pre-trained Temporal Segment Network~\cite{wang2016temporal} (TSN) and extract hand and object RoIAlign~\cite{he2017mask} feature $\mathcal{P}_i^t$ 
given the detected bounding boxes from~\cite{shan2020understanding}. The global feature $\mathcal{P}_g^t$ is obtained similarly by average-pooling. 
Next, for hands and objects, we concatenate the pooled features along with the corresponding center coordinates and forward it to a Multi-Layer Perceptron (MLP), yielding the $X_i^t$. 
Take a hand as an example, $X_h^t = W_h[h_t;\mathcal{P}_h^t]$, $W_h$ is the learnable weights of the corresponding hand MLP. When there is no hand or object detected in a certain frame, we set corresponding places with zero vectors. For global features, we directly use an MLP to obtain the output, $X_g^t=W_g\mathcal{P}_g^t$. All the features (global/hand/object) are taken as independent input tokens to the Transformer. 

\begin{figure*}[t]
    \centering
    \includegraphics[width=0.87\linewidth]{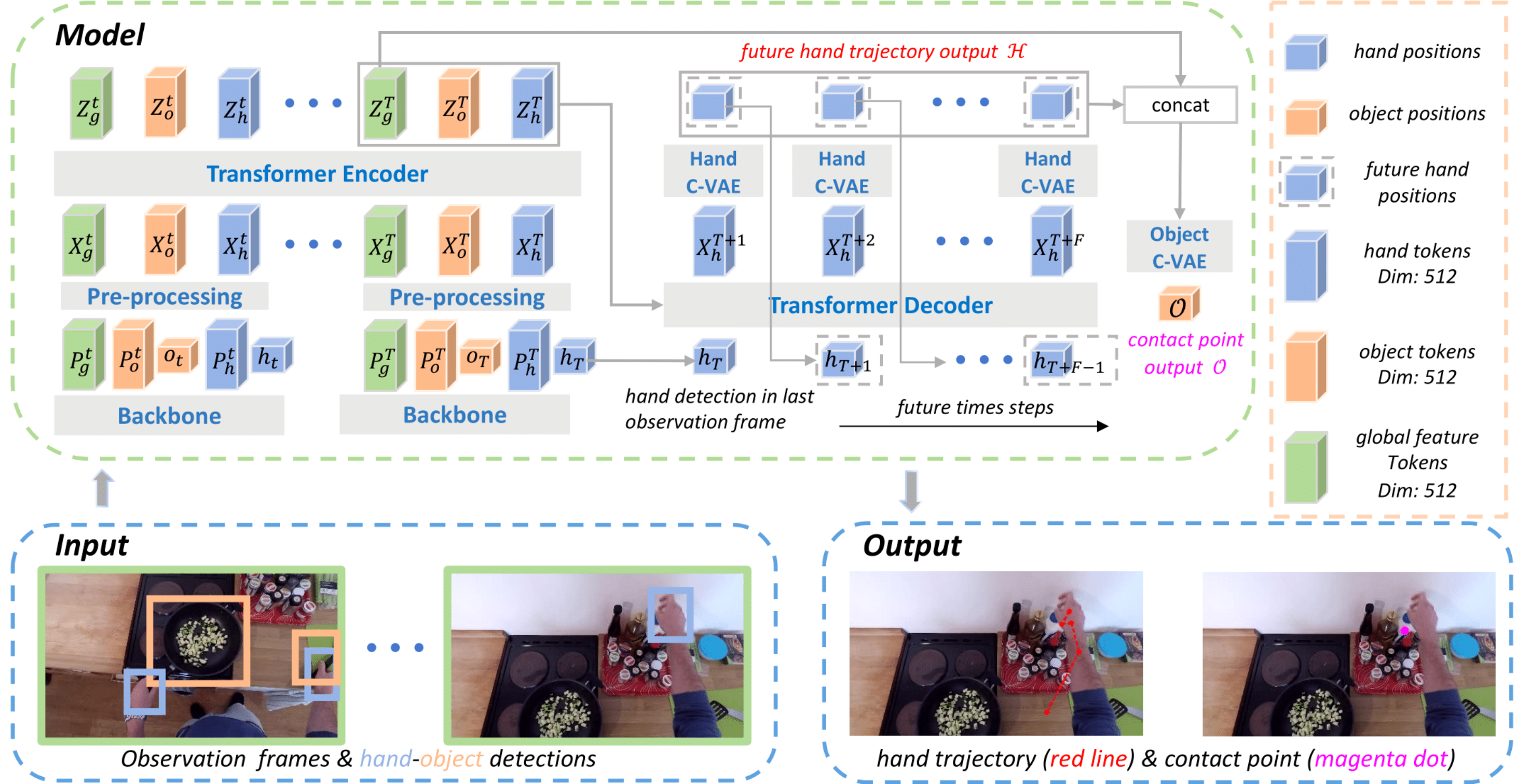}
    \caption{The OCT has an encoder-decoder architecture. The input is observation frames and corresponding hand-object detections. The output is the future hand trajectory and contact point prediction. Inside the model, we use ConvNet to extract hand, object and global features of each frame as input tokens to the Transformer encoder. All tokens (global/hand/object) are passed through the Transformer independently. We take the output from the encoder and previously predicted hand locations as input to the decoder. The decoder output is sent to hand C-VAE and object C-VAE to obtain the final results.}
    \vspace{-0.1in}
    \label{fig: model}
    
\end{figure*}

\section{Object-Centric Transformer}
The proposed Object-Centric Transformer (OCT) has an encoder-decoder architecture as shown in Figure~\ref{fig: model}. 
Both the encoder and the decoder stack multiple basic encoding and decoding blocks $\mathcal{B}$. Each block has an attention module named $\mathrm{Att}$ and a feed-forward module that consists of a two-layer $\mathrm{MLP}$ followed by a layer normalization~\cite{ba2016layer} ($\mathrm{LN}$). The only difference between the two blocks is the attention module, where we perform self-attention across input tokens in the encoding block and cross-attention between the encoder output and prediction in the decoding block. Assume $Q^{\ell-1}$ is the output query from block $\ell-1$, and the key, value, mask denoted by $K^{\ell-1},V^{\ell-1},M$ respectively, are three additional inputs to block $\ell$. Then the output $Q^{\ell}=\mathcal{B}(Q^{\ell-1};K^{\ell-1};V^{\ell-1}, M)$, of the block $\mathcal{B}$ could be written as follows: 
\begin{align*}
& [Q;K;V] = W [Q^{\ell-1};K^{\ell-1};V^{\ell-1}] \\
& Q' = Q^{\ell-1} + \mathrm{Att}(Q, K, V, M) \\
& Q^{\ell} = Q' + \mathrm{MLP}(\mathrm{LN}(Q'))
\end{align*}
where the input tokens $Q^{\ell-1}, K^{\ell-1},V^{\ell-1}$ are first passed through a linear transformation layer parameterized by $W$ to produce embeddings $Q, K, V$; then they are forwarded to the attention module $\mathrm{Att}$. The attention output is sent to the feed-forward module with residual connection~\cite{he2016deep} to obtain the final output $Q^{\ell}$. The attention operator is defined as follows:
\begin{align*}
\mathrm{Att}(Q, K, V, M) = \mathrm{softmax}(\frac{QK^T+M}{\sqrt{D}})V
\end{align*}
$D$ is the dimension of the attention module. The attention operator computes a weighted sum of value $V$ where the weight is computed by the taking dot-product between query $Q$ and key $K$ and adding up the mask $M$ followed by scaling and Softmax normalization. $M$ masks out the padding values in key $K$ by setting corresponding positions to \emph{-inf} before the Softmax calculation.

\subsection{Encoder}
\label{sec: encoder}
The Encoder $\mathcal{E}$ stacks multiple encoding blocks $\mathcal{B}$ and generate outputs $\{Z_1, \cdots, Z_T\}$ from inputs $\{X_1, \cdots, X_T\}$ (Sec.~\ref{sec: preprocess}):
\begin{equation}
    Z_1, \cdots, Z_T = \mathcal{E}(X_1, \cdots, X_T)
\end{equation}
Each $X_t=(X_h^t, X_o^t, X_g^t)$ and $Z_t=(Z_h^t, Z_o^t, Z_g^t)$ consists of a triplet of tokens, hand tokens $h$, object tokens $o$ and global tokens $g$. The input tokens are encoded by two kinds of embeddings, a learnable spatial embedding~\cite{dosovitskiy2020image} to represent category-specific (global/hand/object) information of different features, and the sinusoidal positional embedding~\cite{vaswani2017attention} to incorporate the temporal position information. All tokens passed through the encoding blocks independently. In each encoding block, we compute self-attention over all input tokens across space and time. Given that there could be padded tokens when there is no hand/object detected in certain frames, we use mask $M$ to mask out such tokens. Thus we have $Q^{\ell}=\mathcal{B}(Q^{\ell-1}, Q^{\ell-1}, Q^{\ell-1}, M)$ in each of the $\ell$-th encoding blocks, where the query, key, and value come from the same output $Q^{\ell-1}$. 

\subsection{Decoder}
\label{sec: decoder}
The decoder $\mathcal{D}$ predicts future hand feature $X_{T+t}$ one at a time, where $t \in (T+1, T+F)$ is the future time step. The predicted features $X_{T+t}$ are then sent to the trajectory head network to predict the future hand location $h_{T+t}$. At each step, the decoder is auto-regressive~\cite{graves2013generating}, consuming the previously generated future hand locations $(h_{T}, \cdots, (h_{T+t-1})$ as additional input when generating $X_{T+t}$. The $0$-th input to the decoder is the hand location $h_T$ in the last observation frame. The prediction at future time step $t$ of the decoder can be written as follows:
\begin{equation}
    X_{T+t} = \mathcal{D}(h_T, \cdots, h_{T+t-1})
\end{equation}
The decoder consists of several decoding blocks $\mathcal{B}$. Each block works like an encoding block, except it performs cross-attention that takes the output $Q^{\ell-1}$ from block $\ell-1$ as query, output token $Z_T$ (Sec.~\ref{sec: encoder}) of the encoder in the last observation frame as key and value. To constrain the decoder block to only attend to earlier input positions of $Q^{\ell-1}$, we create a mask $M'$ that masks out subsequent positions. Thus we have $Q^{\ell}=\mathcal{B}(Q^{\ell-1}, Z_T, Z_T, M')$ in each decoding block $\ell$, where the three inputs to block $\mathcal{B}$ correspond to query, key, and value. Before forwarding the input to the 1st decoding block, we encode it with sinusoidal positional embedding~\cite{vaswani2017attention} to incorporate the temporal information. 

\subsection{Head Networks}
We employ two C-VAE as two heads; one for hand trajectory estimation and another for object contact points prediction. 
A C-VAE contains two functions: an encoding function $\mathcal{F}_{enc}$ which encodes the input $x$ and condition $c$ into a latent z-space parameterized by mean $\mu$ and co-variance $\sigma$, and a decoding function $\mathcal{F}_{dec}$ which decodes sampled $z$ from the latent space and condition $c$ to reconstruct input $x$. Formally, we have $\mu, \sigma = \mathcal{F}_{enc}(x; c)$ and $\hat{x} = \mathcal{F}_{dec}(z; c)$ where $z \sim \mathcal{N}(\mu, \sigma^2)$. The $\mathcal{F}_{enc}$ and $\mathcal{F}_{dec}$ are implemented as a MLP. At training time, we minimize the objective of reconstruction error $\mathcal{L}_{recon}(x, \hat{x})=\|x-\hat{x}\|^2$ between ground-truth $x$ and predicted $\hat{x}$, as well as a KL-Divergence term $\mathcal{L}_{kl}(\mu, \sigma)=-KL[\mathcal{N}(\mu, \sigma^2)||\mathcal{N}(0, 1)]$ that regularizes the latent z-space close to normal distribution $\mathcal{N}(0, 1)$. 
During inference, we sample $z$ from the latent space and concatenate with condition $c$ to predict output $\hat{x}$.

\paragraph{Hand C-VAE.} At future time step $t$, the hand C-VAE takes hand locations $h_{T+t}$ as input, and conditioned on the hand feature tokens $X_{T+t}$ (Sec.~\ref{sec: decoder}) from the decoder output. The encoding function $\mathcal{F}_{enc}^h$ outputs distribution parameters $\mu_h$ and $\sigma_h$ of the latent space. The decoding function $\mathcal{F}_{dec}^h$ predicts future hand locations $\hat{h}_{T+t}$. Thus, the loss function of hand C-VAE $\mathcal{L}_{\mathcal{H}}$ is the reconstruction loss over all future time steps $t$ and KL-Divergence regularization:
\begin{equation}
\mathcal{L}_{\mathcal{H}}=\sum_{t=1}^F\mathcal{L}_{recon}(h_{T+t}, \hat{h}_{T+t}) + \mathcal{L}_{kl}(\mu_h, \sigma_h)   
\end{equation}

\paragraph{Object C-VAE.} The object C-VAE takes the future contact points $o$ sampled from the generated ground-truth set of future contact points $\mathcal{O}$ (Sec~\ref{sec: task}) as input, and is conditioned on global feature token $Z_g^T$ (Sec.~\ref{sec: encoder}) in the last observation frame from the Transformer encoder output and future hand locations $(h_T, \cdots, h_{T+F})$. The hand trajectory is forwarded to a fully-connected layer and concatenated with $Z_g^T$ as the conditional input. We found that the object's future contact points could be predicted more accurately with the future hand trajectory as conditional input. During training, we use teacher forcing~\cite{williams1989learning} by taking ground-truth future hand trajectory as input. During inference, we use the predicted future hand trajectory as input for the object C-VAE. Similar to hand C-VAE, the encoding function $\mathcal{F}_{enc}^o$ outputs $\mu_o$ and $\sigma_o$, while the decoding function $\mathcal{F}_{dec}^o$ predicts future object contact points $\hat{h}_{o}$. The loss function, $\mathcal{L}_{\mathcal{O}}$, of the object C-VAE  is the following:
\begin{equation}
\mathcal{L}_{\mathcal{O}}=\mathcal{L}_{recon}(o, \hat{o}) + \mathcal{L}_{kl}(\mu_o, \sigma_o)   
\end{equation}

\subsection{Training and Inference}
\paragraph{Training.}
We train the Object-Centric Transformer with both the hand trajectory loss $\mathcal{L}_{\mathcal{H}}$ and object contact point loss $\mathcal{L}_{\mathcal{O}}$. 
We observe that the object contact points labels are noisier than the hand trajectory labels in our generated training set.
The total loss is $\mathcal{L} = \mathcal{L}_{\mathcal{H}} + \lambda \mathcal{L}_{\mathcal{O}}$, where $\lambda=1e^{-1}$, is a constant coefficient to balance the training loss.

\paragraph{Inference.}
During inference, 
we sample $20$ times for both the trajectories and contact points from the C-VAE for each input video. Following the evaluation protocol in previous work~\cite{ye2019compositional, mangalam2020not, qi2020learning, mohamed2020social} that involves stochastic unit in trajectory estimation, we report the minimum of among $20$ samples for trajectory evaluation. We collect all predicted contact points and convert them into a heatmap by centering a Gaussian distribution over each point for affordance evaluation.

\section{Experiments}

\subsection{Implementation Details}
We sample $T=10$ frames at $4$ FPS (frames per second) as input observations and forecast $1$s in the future on Epic-Kitchens, where the future time horizon $F=4$. We sample $T=9$ frames at $6$ FPS on EGTEA Gaze+, forecasting $0.5$s with $F=3$. We use the pre-trained TSN~\cite{wang2016temporal} from~\cite{furnari2020rolling} as the backbone to extract RGB features from the input video clip. We use the detector proposed in~\cite{shan2020understanding} to detect active hand and object bounding boxes in each input frame. Then we use RoIAlign~\cite{he2017mask} and average pooling to produce a $1024$-D vector for hand $\mathcal{P}_h^t$, object $\mathcal{P}_o^t$ and global features $\mathcal{P}_g^t$ (Sec.~\ref{sec: preprocess}) at input time step $t$. We set the embedding dimension of the OCT to $512$. We set the number of blocks in encoder and decoder to be $6$ and $4$ on Epic-Kitchens, $2$ and $1$ on EGTEA Gaze+. Each block has $8$ attention heads. For encoding and decoding function $\mathcal{F}_{enc}$ and $\mathcal{F}_{dec}$ in C-VAE, we use a single-layer MLP for both the hand and object. The OCT is trained using Adam optimizer~\cite{kingma2014adam} with a learning rate of $1e-4$ and a batch size of $128$. Training takes $35$ epochs on Epic-Kitchens, $25$ epochs on EGTEA Gaze+, including $5$ epochs warm-up~\cite{goyal2017accurate} and rest epochs with cosine decay~\cite{loshchilov2016sgdr}. During inference, we sample $20$ times from the C-VAE for both hand trajectory and object contact points. Please see supplementary for detailed network structures.

\subsection{Datasets}
We use Epic-Kitchens-55 (EK55)~\cite{damen2018scaling}, Epic-Kitchens-100 (EK100)~\cite{damen2021rescaling} and EGTEA Gaze+ (EG)~\cite{li2018eye} datasets for experiments. The EK100 dataset is an extended version of the EK55 dataset. All datasets capture daily activities in the kitchen. Following the standard partition protocol in~\cite{furnari2020rolling, damen2021rescaling}, we split the training set of both datasets into training and validation splits. Given the test set are only used for action anticipation, we don't incorporate them in our experiments. We used the method in Sec.~\ref{sec: generation} to generate training labels automatically. The evaluation is performed on the validation split of all datasets. We manually filtered out badly generated hand trajectories and collected interaction hotspot annotations on a challenging subset via the Amazon Mechanical Turk platform (see supplementary for details). Given the last observation frame and contact frame in the future, we ask workers to place 1-5 future contact points in the last observation frame. Following~\cite{Fang_2018_CVPR, nagarajan2019grounded}, we convert these annotations into an affordance heatmap as our ground-truth. On the EK55 dataset, we collect $8523$ training samples, $1894$ evaluation hand trajectories, and $241$ interaction evaluation hotspots. On the EK100 dataset, we collect $24148$ training samples, $3513$ evaluation hand trajectories, and $401$ evaluation interaction hotspots. On the EG dataset, we collect $1880$ training samples, $442$ evaluation hand trajectories, and $69$ evaluation interaction hotspots.

\subsection{Evaluation Metrics}
\paragraph{Trajectory evaluation.} We use normalized predicted 2D hand locations for evaluation using the following metrics.  
\begin{itemize}
    \item \textbf{Average Displacement Error (ADE)}. ADE is calculated as the $\ell_2$ distance between the predicted future and the ground-truth averaged over the entire trajectory and two hands. 
    \item \textbf{Final Displacement Error (FDE)}. FDE measures the $\ell_2$ distance between the predicted future and ground truth at the last time step and averaged over two hands. 
\end{itemize}

\paragraph{Interaction hotspots evaluation.} We downsample and normalize the affordance heatmap with a resolution of $32x$ and ensure it sums up to $1$. We don't use KLD (Kullback-Leibler Divergence) metric~\cite{bylinskii2018different} as it is known to be sensitive to the tail of the distributions~\cite{zhang2010simplifying, ben2015kullback, ozair2019wasserstein}. A small difference in the low-density regions may induce a huge KLD, especially severe for forecasting problems.  

\begin{itemize}
    \item \textbf{Similarity Metric (SIM)}: SIM~\cite{swain1991color} measures the similarity between the predicted affordance map distribution and the ground-truth one. It is computed as the sum of the minimum values at each pixel location between the predicted map and the ground-truth map. 
    \item \textbf{AUC-Judd (AUC-J)}: AUC-J~\cite{judd2009learning} is a variant of AUC proposed by Judd \textit{et al.}~\cite{judd2009learning}. The AUC evaluates the ratio of ground-truth captured by the predicted affordance map under different thresholds~\cite{bylinskii2018different}. 
    \item \textbf{Normalized Scanpath Saliency (NSS)}: NSS~\cite{peters2005components} measures the correspondence between the predicted affordance map and the ground truth. It is computed by normalizing the predicted affordance map to have zero mean and unit standard deviation and averaging over ground truth locations. 
\end{itemize}

\subsection{Comparison to the state-of-the-art}

\begin{table}[t]
\caption{\textbf{Future hand trajectory estimation performance} on three datasets. ($\uparrow$/$\downarrow$ indicates higher/lower is better.) Our method outperforms previous approaches by a large margin and achieves comparable performance with the more elaborate divided space-time attention design.}
\label{tab: exp-traj}
\centering
\setlength{\tabcolsep}{10pt}
\resizebox{1\linewidth}{!}{
\begin{tabular}{c cc cc cc}
\toprule
\multicolumn{1}{c}{}  & \multicolumn{2}{c}{EK55} & \multicolumn{2}{c}{EK100} & \multicolumn{2}{c}{EG}\\
\cmidrule(lr){2-3} \cmidrule(lr){4-5} \cmidrule(lr){6-7}
Methods & ADE  $\downarrow$ & FDE  $\downarrow$ & ADE  $\downarrow$ & FDE  $\downarrow$ & ADE  $\downarrow$ & FDE $\downarrow$\\
\shline
KF~\cite{bewley2016simple} & 0.34 & 0.33 & 0.33 & 0.32 & 0.49 & 0.48 \\
Seq2Seq~\cite{sutskever2014sequence} & 0.18 & 0.14 & 0.18 & 0.14 & 0.18 & 0.14\\
FHOI~\cite{liu2020forecasting} & 0.36 & 0.35 & 0.35 & 0.35 & 0.34 & 0.34\\
\midrule
Divided & \textbf{0.11} & \textbf{0.11} & \textbf{0.12} & \textbf{0.11} & 0.15 & 0.15\\
Ours & 0.12 & 0.12 & \textbf{0.12} & \textbf{0.11} & \textbf{0.14} & \textbf{0.14}\\
\bottomrule
\end{tabular}
}
\end{table}

\paragraph{Trajectory estimation.} We evaluate our method against several baselines and state-of-the-art approaches. Kalman Filter (\textbf{KF})~\cite{bewley2016simple} tracks the center of the hand in observation frames and predicts future hands locations. \textbf{Seq2Seq}~\cite{sutskever2014sequence} used LSTM to encode temporal information in the observation sequence and decode the target locations. Forecasting HOI (\textbf{FHOI})~\cite{liu2020forecasting} used I3D~\cite{carreira2017quo} (CNN) with motor attention to forecast future hand motion. Note that \textbf{FHOI} only used observation frames as input without accessing to hand-object detections. Besides, we also compare against Divided Attention~(\textbf{Divided})~\cite{bertasius2021space} Transformer design by applying temporal attention and spatial attention separately in the encoder of the OCT instead of doing them jointly (Sec.~\ref{sec: encoder}). We compute temporal attention only across hand tokens in different frames and spatial attention only within each frame. The results are shown in Table~\ref{tab: exp-traj}. Experimental results show that our method outperforms previous approaches by a large margin, improving the ADE by $50\%$, and FDE by $27.3\%$ on the EK100 dataset against the second-best method of each metric, and achieves similar performance with the Divided Attention Transformer encoder design. This demonstrates the superiority of using Transformer to capture hand, object, and environment context interactions in egocentric videos.

\begin{table}[t]
\caption{\textbf{Future object interaction hotspots prediction performance} on three datasets. ($\uparrow$/$\downarrow$ indicates higher/lower is better.) Our method outperforms prior work as well as Divided Attention significantly.
}
\label{tab: exp-obj}
\centering
\setlength{\tabcolsep}{3pt}
\resizebox{1\linewidth}{!}{
\begin{tabular}{c ccc ccc ccc}
\toprule
\multicolumn{1}{c}{}  & \multicolumn{3}{c}{EK55} & \multicolumn{3}{c}{EK100} & \multicolumn{3}{c}{EG}\\
\cmidrule(lr){2-4} \cmidrule(lr){5-7} \cmidrule(lr){8-10}
Methods & SIM $\uparrow$ & AUC-J $\uparrow$ & NSS $\uparrow$ & SIM $\uparrow$ & AUC-J$\uparrow$ & NSS $\uparrow$ & SIM $\uparrow$ & AUC-J$\uparrow$ & NSS $\uparrow$\\
\shline
Center & 0.09 & 0.61 & 0.33 & 0.09 & 0.62 & 0.31 & 0.09 & 0.63 & 0.27\\
Hotspots~\cite{nagarajan2019grounded} & 0.15 & 0.66 & 0.53 & 0.14 & 0.66 & 0.47 & 0.15 & 0.71 & 0.69\\
FHOI~\cite{liu2020forecasting} & 0.13 & 0.57 & 0.21 & 0.12 & 0.56 & 0.18 & 0.15 & 0.66 & 0.51\\
\midrule
Divided & 0.19 & 0.67 & 0.67 & 0.16 & 0.66 & 0.50 & 0.19 & 0.70 & 0.69\\
Ours & \textbf{0.22} & \textbf{0.70} & \textbf{0.87} & \textbf{0.19} & \textbf{0.69} & \textbf{0.72} & \textbf{0.23} & \textbf{0.75} & \textbf{1.01} \\
\bottomrule
\end{tabular}
}
\end{table}

\begin{table}[t]
\centering
\begin{minipage}{0.23\textwidth}
\caption{\textbf{Cross-dataset hand trajectory estimation generalization performance}. All models are trained on Epic-Kitchens and tested on EGTEA Gaze+.}
    \label{tab: exp-traj-cross}
    \footnotesize
    \setlength{\tabcolsep}{2.5pt}
    \centering
    \begin{tabular}{l cc}
    \toprule
    Methods & ADE  $\downarrow$ & FDE  $\downarrow$ \\
    \shline
    Seq2Seq~\cite{sutskever2014sequence} & 0.24 & 0.19 \\
    FHOI~\cite{liu2020forecasting} & 0.31 & 0.32\\
    \midrule
    Divided & \textbf{0.15} & \textbf{0.13} \\
    Ours & 0.16 & \textbf{0.13} \\
    \bottomrule
    \end{tabular}
\end{minipage}
\hfill
\begin{minipage}{0.23\textwidth}
\caption{\textbf{Cross-dataset interaction hotspots prediction generalization performance}. All models are trained on Epic-Kitchens and tested on EGTEA Gaze+.}
    \label{tab: exp-obj-cross}
    \footnotesize
    \setlength{\tabcolsep}{2.5pt}
    \centering
    \begin{tabular}{l ccc}
    \toprule
    Methods & SIM $\uparrow$ & AUC-J$\uparrow$ & NCC $\uparrow$\\
    \shline
    Hotspots~\cite{nagarajan2019grounded}  & 0.15 & 0.71 & 0.69 \\
    FHOI~\cite{liu2020forecasting} & 0.12 & 0.54 & 0.10 \\
    \midrule
    Divided & 0.21 & 0.74 & 0.80 \\
    Ours & \textbf{0.23} & \textbf{0.78} &  \textbf{1.02} \\
    \bottomrule
    \end{tabular}
\end{minipage}
\vspace{-0.1in}
\end{table}

\paragraph{Interaction hotspots prediction.} We compare our results with the following methods. \textbf{Center}~\cite{nagarajan2019grounded, liu2020forecasting, luo2021learning} generated the heatmap by placing a fixed Gaussian at the center of the image. \textbf{Hotspots}~\cite{nagarajan2019grounded} anticipated spatial interaction regions using Grad-Cam~\cite{selvaraju2017grad}, given the future action label as additional input. 
\textbf{FHOI}~\cite{liu2020forecasting} and \textbf{Divided}~\cite{bertasius2021space} are the same method and baseline introduced in trajectory estimation, where they used I3D (CNN) and divided space-time Transformer encoder respectively. Table~\ref{tab: exp-obj} summarizes the results of interaction hotspots prediction. Our method achieves the best performance across datasets and all metrics, improves SIM by $+5\%$, AUC-J by $+3\%$, and NSS by $+25\%$ on the EK100 dataset against the second-best method of each metric. Compared to Divided Attention, jointly modeling all hand-object tokens in observation frames is more beneficial for the prediction. These results also highlight that the Transformer architecture is more suitable for visual forecasting problems.

\paragraph{Cross dataset generalization.} We evaluate learned models' cross-dataset generalization ability on both tasks. All models are trained on Epic-Kitchens and tested on EGTEA Gaze+. The hand trajectory estimation and interaction hotspots prediction performances are shown in Table~\ref{tab: exp-traj-cross} and Table~\ref{tab: exp-obj-cross} respectively. In addition to superior in-domain performance, our method demonstrates strong cross-domain generalization by significantly outperforming other approaches across all metrics on both tasks.

\subsection{Ablations and Analysis}
We do ablation studies of our method on the EK100 dataset.

\begin{table}[t]
\centering
\begin{minipage}{0.23\textwidth}
    \caption{Ablation study of \textbf{trajectory estimation by using different head network}. Stochastic models are in \textbf{bold}.}
    \label{tab: exp-traj-head}
    \footnotesize
    \setlength{\tabcolsep}{2.5pt}
    \centering
    \begin{tabular}{l cc}
    \toprule
    Heads & ADE  $\downarrow$ & FDE  $\downarrow$ \\
    \shline
    MLP & 0.21 & 0.16\\
    \textbf{Bivariate} & 0.19 & 0.14 \\
    \textbf{C-VAE} & \textbf{0.12} & \textbf{0.11}\\
    \bottomrule
    \end{tabular}
\end{minipage}
\hfill
\begin{minipage}{0.23\textwidth}
    \caption{Ablation study of \textbf{hotspots prediction by using different head network}. Stochastic models are in \textbf{bold}.}
    \label{tab: exp-obj-head}
    \footnotesize
    \setlength{\tabcolsep}{2.5pt}
    \centering
    \begin{tabular}{l ccc}
    \toprule
    Heads & SIM $\uparrow$ & AUC-J$\uparrow$ & NCC $\uparrow$\\
    \shline
    MLP & 0.14 & 0.59 & 0.43\\
    \textbf{MDN} & 0.16 & 0.64 & 0.53\\
    \textbf{C-VAE} & \textbf{0.19} & \textbf{0.69} & \textbf{0.72}\\
    \bottomrule
    \end{tabular}
\end{minipage}
\vspace{-0.1in}
\end{table}

\paragraph{Head ablation.} 
First, we evaluate the performance of using different stochastic/deterministic head networks for trajectory estimation and contact points prediction. For trajectory estimation, we compare proposed \textbf{C-VAE} with \textbf{MLP} and \textbf{Bivariate}. MLP deterministically outputs the future hand locations, while the Bivariate~\cite{alahi2016social} assumes the future hand location follows a bivariate Gaussian distribution at each time step and explicitly samples from the predicted distribution during inference. For future contact points prediction, we compare \textbf{C-VAE} with \textbf{MLP} and \textbf{MDN}. MDN~\cite{bishop1994mixture} adopts the Mixture Density Model (MDN) and models the distribution of future contact points as a mixture of Gaussians, where we set the number of Gaussian components to be $3$. As shown in Table~\ref{tab: exp-traj-head} and Table~\ref{tab: exp-obj-head}, stochastic models outperform the deterministic one on both tasks, thanks to their ability to deal with uncertainty. 
Adopting C-VAE against MLP improves the trajectory estimation performance by $75.0\%$ on ADE and $45.5\%$ on FDE, also obtains $+5\%$, $+10\%$, and $+29\%$ gain on SIM, AUC-J, and NCC of hotspots prediction. Besides, we also observe that C-VAE achieves better results compared to Bivariate and MDN. It demonstrates modeling stochastic in latent space works better than output space.

\begin{table}[t]
\centering
\caption{Ablation study of \textbf{different C-VAE conditions}. $\mathcal{H}$ and $\mathcal{O}$ are future hand trajectory and contact point.  $\mathcal{O}|\mathcal{H}$ stands for object C-VAE is conditioned on hand trajectory, similar for $\mathcal{H}|\mathcal{O}$. None means no conditions. Predicting contact points conditioned on hand trajectory gives the best performance for both tasks.}
\label{tab: exp-cvae}
\footnotesize
\setlength{\tabcolsep}{5pt}
\begin{tabular}{c cc ccc}
\toprule
\multicolumn{1}{c}{} & \multicolumn{2}{c}{Trajectory} & \multicolumn{3}{c}{Interaction Hotspots}\\
\cmidrule(lr){2-3} \cmidrule(lr){4-6}
Condition & ADE  $\downarrow$ & FDE  $\downarrow$ & SIM $\uparrow$ & AUC-J$\uparrow$ & NCC $\uparrow$ \\
\shline
None & 0.14 & 0.12 & 0.16 & 0.64 & 0.53\\
$\mathcal{H}|\mathcal{O}$ & 0.13 & 0.12 & 0.16 & 0.64 & 0.54 \\
$\mathcal{O}|\mathcal{H}$ & \textbf{0.12} & \textbf{0.11} & \textbf{0.19} & \textbf{0.69} & \textbf{0.72}\\
\bottomrule
\end{tabular}
\end{table}

\paragraph{C-VAE condition.} Besides modeling uncertainty in C-VAE, we analyze the effect of condition dependency in C-VAE. In Table~\ref{tab: exp-cvae}, we evaluate the performance of using different C-VAE conditions for both the hand and the object. We compare three cases: no condition between the hand trajectory and object contact point, denoted as \textbf{None}; hand trajectory is conditioned on object contact point, denoted as \bm{$\mathcal{H}|\mathcal{O}$}; object contact points is conditioned on hand trajectory, denoted as \bm{$\mathcal{O}|\mathcal{H}$}. We find that explicitly incorporating the conditional dependency in C-VAE improves the overall performance. Predicting interaction hotspots conditioned on future hand trajectory leads to the best result on both tasks, obtaining $+3\%$, $+5\%$, and $+18\%$ performance gain on SIM, AUC-J, and NCC against conditioned on the inverse order. It suggests that the two tasks are intertwined and modeling their relation explicitly benefits the performance.

\begin{table}[t]
\centering
\caption{Ablation study of \textbf{leveraging more automatically annotated training data}. We compare two models trained on EK55 and KE100 training split under the same setting, and evaluate the performance on EK100 validation split. Training with more automatically annotated training data (EK100) gives better performance on both tasks.}
\label{tab: exp-data}
\footnotesize
\setlength{\tabcolsep}{3pt}
\resizebox{1\linewidth}{!}{
\begin{tabular}{cc cc ccc}
\toprule
\multicolumn{1}{c}{} & \multicolumn{1}{c}{} & \multicolumn{2}{c}{Trajectory} & \multicolumn{3}{c}{Interaction Hotspots} \\
\cmidrule(lr){3-4} \cmidrule(lr){5-7}
Train & Evaluation & ADE $\downarrow$ & FDE  $\downarrow$ & SIM $\uparrow$ & AUC-J$\uparrow$ & NCC $\uparrow$\\
\shline
EK55 & EK100 & 0.13 & 0.12 & 0.18 & 0.68 & 0.60\\
EK100 & EK100 & \textbf{0.12} & \textbf{0.11}  & \textbf{0.19} & \textbf{0.69} & \textbf{0.72}\\
\bottomrule
\end{tabular}
}
\end{table}

\paragraph{More training data.} As we generate our training data automatically without manual labeling, we are interested in understanding whether leveraging more automatically annotated training data can help boost performance. We trained two models under the same setting on EK55 and EK100 training split respectively. We evaluate their performances on the manually-collected EK100 validation split having no overlap with both EK55 and EK100 training splits. As shown in Table~\ref{tab: exp-data}, we observe that a model trained with larger data (EK100) outperforms a model trained on EK55 on both tasks. This demonstrates the effectiveness of our method. Even though there is inevitable noise introduced during training data generation, our method could still learn useful representations for forecasting and it benefits from utilizing more training data. It also indicates a great potential for deploying our method on larger-scale egocentric videos. 

\begin{table}[t]
\begin{minipage}{0.45\linewidth}
    \centering
    \captionof{table}{Ablation study of \textbf{different input features} for trajectory estimation. Global features that encode environmental context and hand features are most important.}
    \label{tab: exp-feat}
    \footnotesize
    \setlength{\tabcolsep}{3pt}
    \resizebox{\linewidth}{!}{
    \begin{tabular}{ccc|cc}
        \toprule
         hand & object & global & ADE  $\downarrow$ & FDE  $\downarrow$\\
         \shline
         \xmark & \cmark & \cmark & 0.13 & 0.16\\
          \cmark & \xmark  & \cmark & 0.13 & \textbf{0.11}\\
         \cmark & \cmark & \xmark & 0.15 & 0.13 \\
         \cmark & \cmark & \cmark & \textbf{0.12} & \textbf{0.11}\\
         \bottomrule
    \end{tabular}}
\end{minipage}
\hfill
\begin{minipage}{0.45\linewidth}
    \centering
    \includegraphics[width=\linewidth]{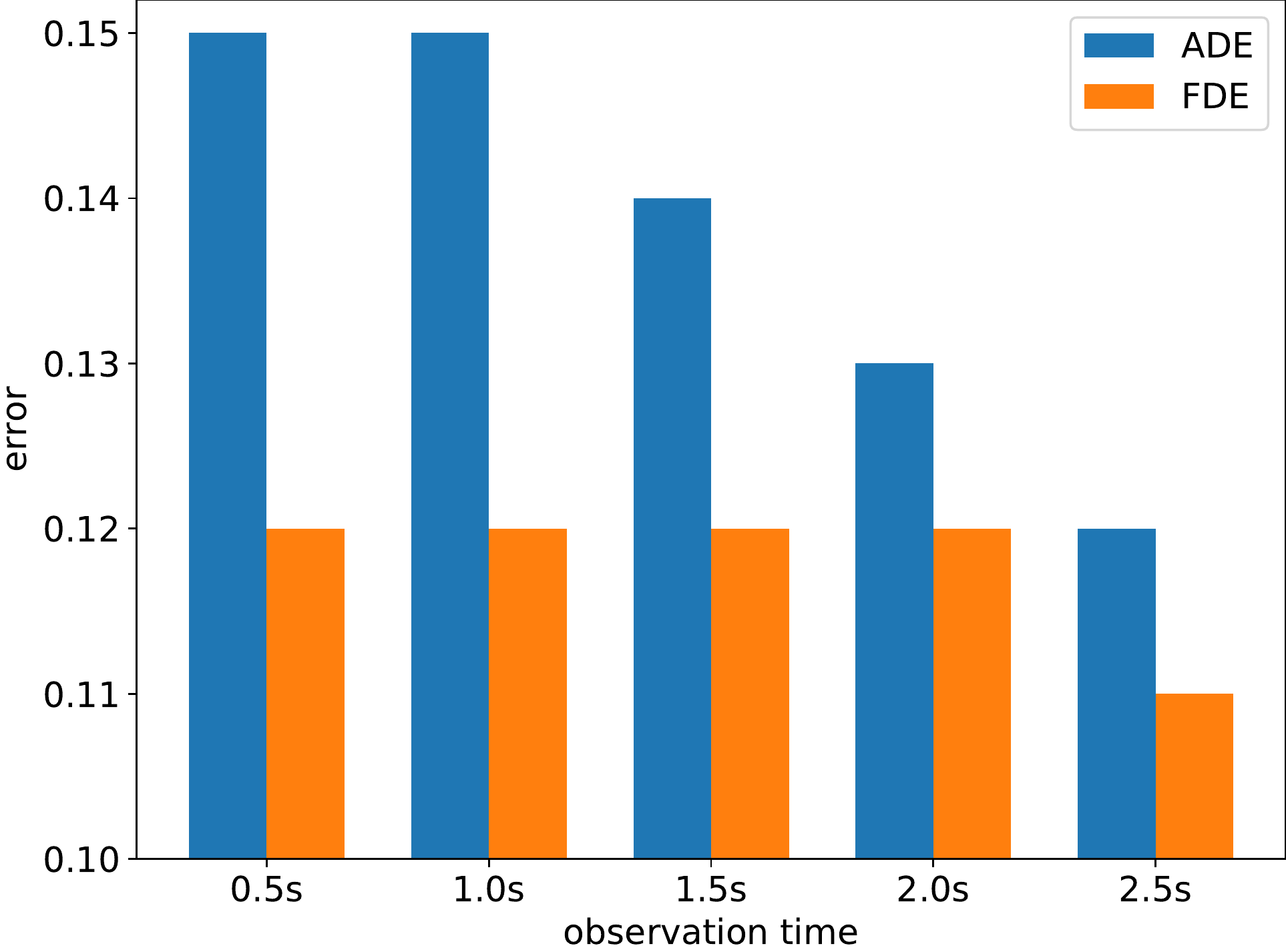}
    \captionof{figure}{Ablation study of \textbf{observation time} contributes to trajectory estimation. Longer temporal context are helpful.}\label{fig: exp-observe}
\end{minipage}
\end{table}

\paragraph{Input ablation.} We evaluate the contributions of different input settings to the performance of trajectory estimation. Contact points prediction performance is conditioned on the trajectory, thus the input relation is not as straightforward as trajectory estimation. We evaluate the contribution of different features by removing them from the input and seeing the performance drop. As shown in Table~\ref{tab: exp-feat}, global features that encode environmental context (Sec.~\ref{sec: preprocess}) and hand features are more crucial to the performance. By removing global features from input, ADE metric drops $25.0\%$. Without hand features as input, FDE metric drops $45.5\%$. This demonstrates that the global features are as imperative as hand features to the trajectory estimation. Besides, we also analyze the effect of different observation lengths in Figure~\ref{fig: exp-observe}. We observe that the performance improves as we incorporate more observation frames as input, which also proves our model is capable of capturing useful temporal information. 

\begin{table}[t]
\centering
\caption{\textbf{Action anticipation performance} on EK55 and EK100 validation split. We report top-5 accuracy/recall in terms of verb, noun, and action on EK55/EK100 respectively,  following~\cite{furnari2019would, furnari2020rolling, girdhar2021anticipative}. We add a single MLP on top of the OCT encoder to classify future actions. We compare the model trained from scratch (\textbf{Scratch}) and the model pre-trained using trajectory and hotspots estimation task (\textbf{Fine-tune}). The fine-tuned model greatly outperforms the model trained from scratch across all anticipation metrics on both datasets.}
\label{tab: exp-ant}
\setlength{\tabcolsep}{5pt}
\footnotesize
\begin{tabular}{l ccc ccc}
\toprule
\multicolumn{1}{c}{} & \multicolumn{3}{c}{EK55} & \multicolumn{3}{c}{EK100} \\
\cmidrule(lr){2-4} \cmidrule(lr){5-7}
Train  & Verb & Noun & Action  & Verb & Noun & Action\\ \shline
Scratch  & 68.7 & 36.1 & 18.9 & 18.9 & 24.0 & 10.0  \\
Fine-tune & \textbf{73.9} & \textbf{45.9} & \textbf{24.4}  & \textbf{21.9} & \textbf{27.6} & \textbf{12.4} \\
\bottomrule
\end{tabular}
\end{table}

\paragraph{Action anticipation.} So far we have shown the capability of our method on the trajectory estimation and interaction hotspots prediction. We further investigate the potential of our trained model for action anticipation task. We only use the OCT encoder and add a single MLP on top of it that takes the output global feature token in the last observation frame $Z_g^T$ (Sec.~\ref{sec: encoder}) from the encoder as an input and predicts future action labels. Following previous work~\cite{furnari2019would, furnari2020rolling, girdhar2021anticipative}, we report top-5 accuracy on EK55 dataset, and top-5 recall on EK100 dataset for verb/noun/action predictions. Each action label consists of \textit{(verb, noun)}. We trained our model on the same training split as we used for trajectory estimation and interaction hotspots prediction and evaluated the performance on corresponding validation splits. We only used cross-entropy loss between the prediction and ground-truth action labels during training. We obtained verb/noun prediction scores by marginalizing over the action scores. We compared two training strategies: training the model from scratch, denoted as \textbf{Scratch}, and fine-tuning the model pre-trained on trajectory and hotspots prediction task, denoted as \textbf{Fine-tune}. In the \textbf{Fine-tune} version, we freeze the OCT encoder and only trained the added MLP. The action anticipation performance of the two methods is shown in Table~\ref{tab: exp-ant}. The \textbf{Fine-tune} model outperforms \textbf{Scratch} model by a large margin across all datasets and all metrics. Specifically, \textbf{Fine-tune} obtains $+4.3\%$, $+8.5\%$, $+4.4\%$ performance gain on EK55 dataset, and $+2.9\%$, $+3.8\%$, $+2.6\%$ performance gain on EK100 dataset. Note that the performance of our model is not fully comparable with state-of-the-art action anticipation models as we only use a subset of samples for training and evaluation. Neither do we adopt any fancy tricks, network structures, or other loss functions for action anticipation. The experimental results show that the representation learned on trajectory estimation and interaction hotspots prediction could benefit the action anticipation task. This also proves the usefulness of the two tasks and generalization to other forecasting tasks. 

\begin{figure}[t]
\centering
\includegraphics[width=\linewidth]{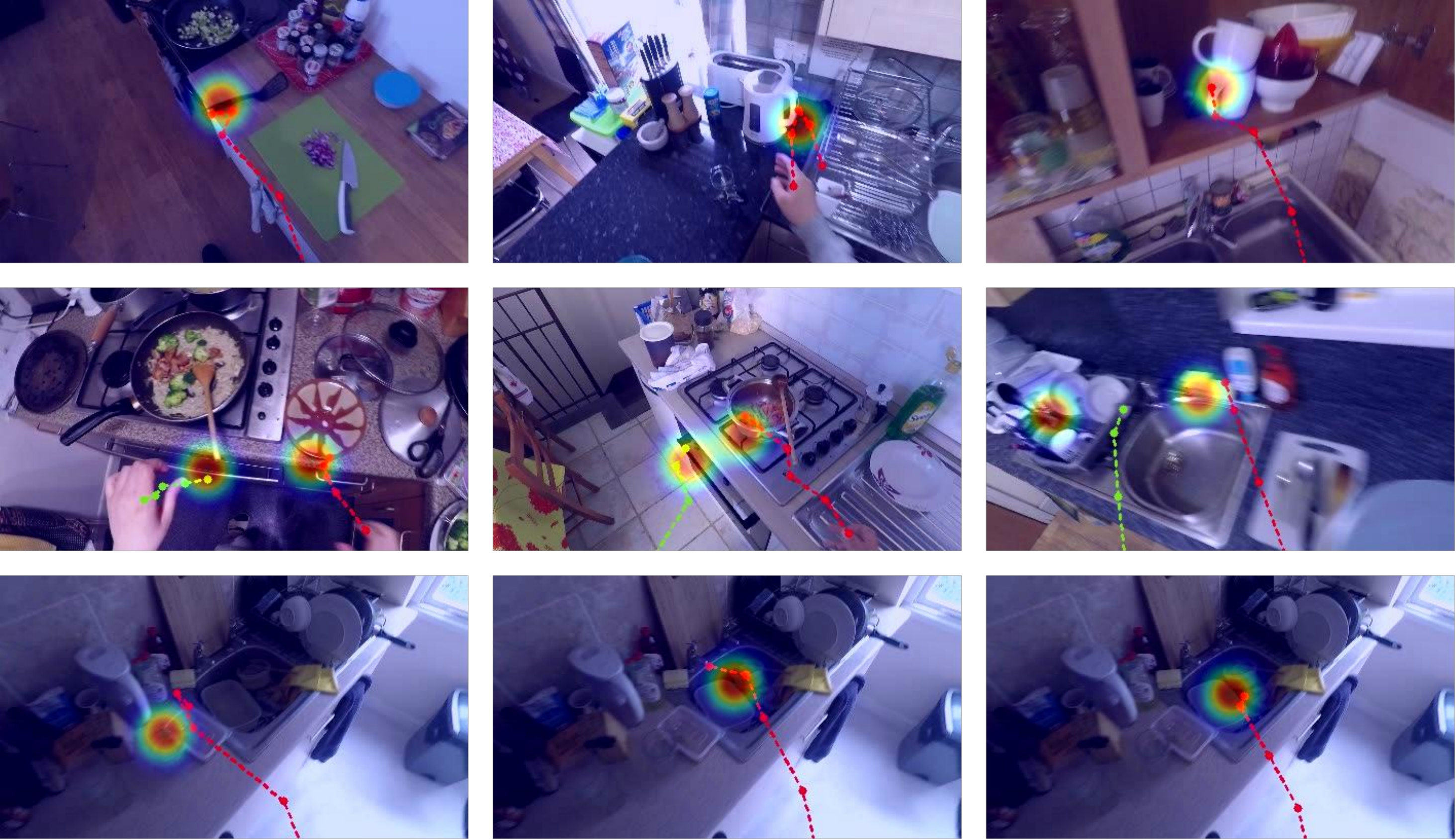}
\caption{Qualitative visualization of future hand trajectory and interaction hotspots. The right and left hand trajectory are shown in \textcolor{red}{red} and \textcolor{green}{green}. The first two row shows single-hand and two-hand scenarios. The third row shows diversity in future trajectory and interaction hotspots prediction.}
\vspace{-0.1in}
\label{fig: vis_predict}
\vspace{-0.1in}
\end{figure}

\paragraph{Qualitative visualization.} 
We visualize predicted future hand trajectory and interaction hotspots in Figure~\ref{fig: vis_predict}. Our method could deal with single and two hands scenarios (when hands are visible in the last observation frame) in the first two rows. Our method can also generate diverse predictions of the future in the third row. This demonstrates that our method is able to forecast the future hand-object interaction considering the future uncertainty. Please see supplementary for more visualizations.

\vspace{-0.05in}
\section{Discussion}
\vspace{-0.05in}

\noindent \textbf{Conclusion.}
We propose to forecast future hand-object interactions in egocentric videos. We solve this task by proposing an automatic way to collect training data, and a novel Object-Centric Transformer (OCT) model that jointly predicts future hand trajectory and interaction hotspots given a sequence of observation frames as input. Through extensive experiments and ablations, we show that OCT significantly outperforms state-of-the-art approaches, and could benefit from stochastic modeling of the future and conditional dependency of trajectory and interaction hotspots into account. Furthermore, we show that our proposed method could leverage more training data to achieve better performance and easily adapt to action anticipation task. In the future, we hope to apply our method to solve more visual forecasting problems in egocentric videos with less human supervision.

\noindent \textbf{Limitation and future Work.}
Our training dataset generation process relies on widely used off-the-shelf tools such as active hand-object detectors and skin segmentation. Thus the ground-truth annotations for training might be affected by the bias and errors from the off-the-shelf tools. In future work, we plan to incorporate self-supervision signals during training to make our model more robust to label noise. 

{\footnotesize \textbf{Acknowledgements.}~Prof. Wang’s lab is supported, in part, by grants from NSF CCF-2112665 (TILOS).}

{\small
\bibliographystyle{ieee_fullname}
\bibliography{egbib}
}

\renewcommand\thefigure{\thesection.\arabic{figure}}
\renewcommand\thetable{\thesection.\arabic{table}}
\setcounter{section}{0}
\setcounter{figure}{0} 
\setcounter{table}{0} 
\appendix

\clearpage
\begin{center}
\textbf{\Large Appendix}
\end{center}

The supplementary material provides more details, results and visualizations to support the main paper. In summary, we include
\begin{itemize}
    \item \ref{supp: training-labels}. Training labels generation details. 
    \item \ref{supp: evaluation}. Evaluation set annotation details.
    \item \ref{supp: implementation}. Implementation details of training and inference. \item \ref{supp: network}. Detailed network architecture and corresponding input and output dimensionality.
    \item \ref{supp: exp}. Additional experiments of comparison Transformer against 3D CNN and adopting end-to-end training.  
    \item \ref{supp: vis-train}. Automatic generated training labels visualization.
    \item \ref{supp: vis}. Qualitative comparison with other methods
    \item \ref{supp: vis-gen}. Cross-environment and cross-dataset generalization results visualization. 
    
\end{itemize}

\section{Training Labels Generation Details}
\label{supp: training-labels}

\paragraph{Hand trajectory generation.} We provide additional implementation details of future hand trajectory training labels generation. As we project hand locations from all future frames to the last observation frame, we need to handle the case when there are missing hand detections in future frames. We fill the gap of missing time steps by conducting Hermite spline interpolation. Such interpolation guarantees the smoothness and continuousness of the generated trajectory. We generate the future hand trajectory at $20$ FPS and sample at $4$ FPS for training.

\begin{figure}[t]
\centering
\includegraphics[width=\linewidth]{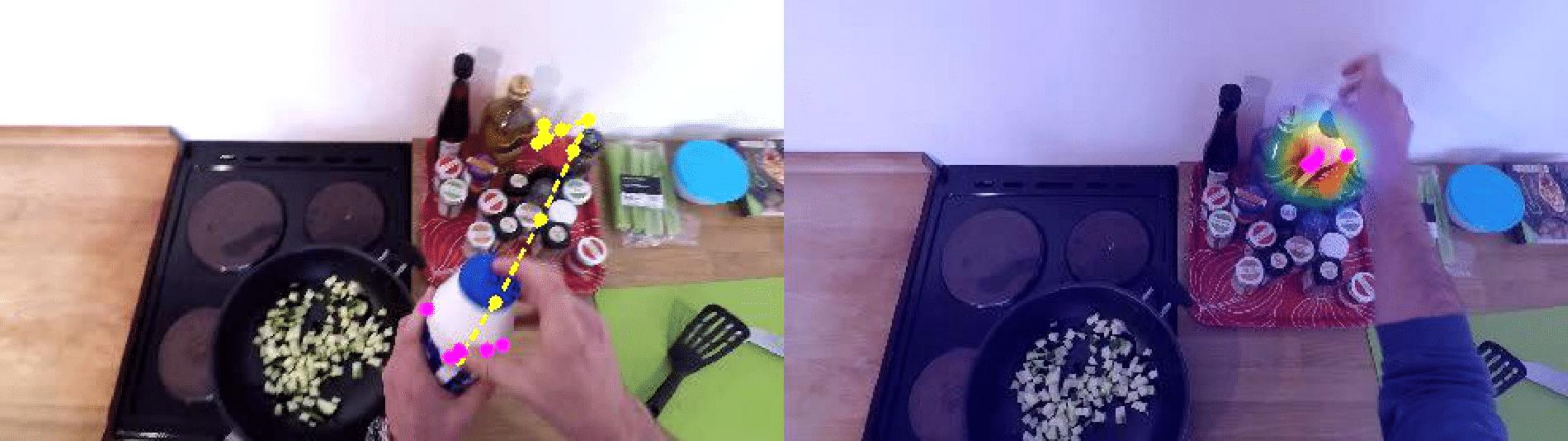}
\caption{Demonstration of how to generate interaction hotspots in active object case. The left image shows the contact frame, while the right image shows the last observation frame. The detected contact points are shown in \textcolor{magenta}{magenta} dots in both frames. We move the detected contact points along the active object future trajectory (\textcolor{yellow}{yellow} line in the left image) to its original place in the last observation frame to compute the correct interaction hotspots.}
\label{fig: vis_affordance}
\end{figure}

\paragraph{Interaction hotspots generation.} For interaction hotspots training label generation, we detect contact points in the contact frame and project them back to the last observation frame by a similar technique as future hand trajectory generation. However, we need to handle the active object case, \textit{i.e.} the object is moved by the hand in future frames, as shown in Figure~\ref{fig: vis_affordance}. To this end, we obtain a future active object trajectory similar to the hand and move the contact points to the active object's original place where it stays still in the last observation frame after the projection.

\section{Evaluation Annotations Details}
\label{supp: evaluation}
\begin{figure}[t]
\centering
\includegraphics[width=\linewidth]{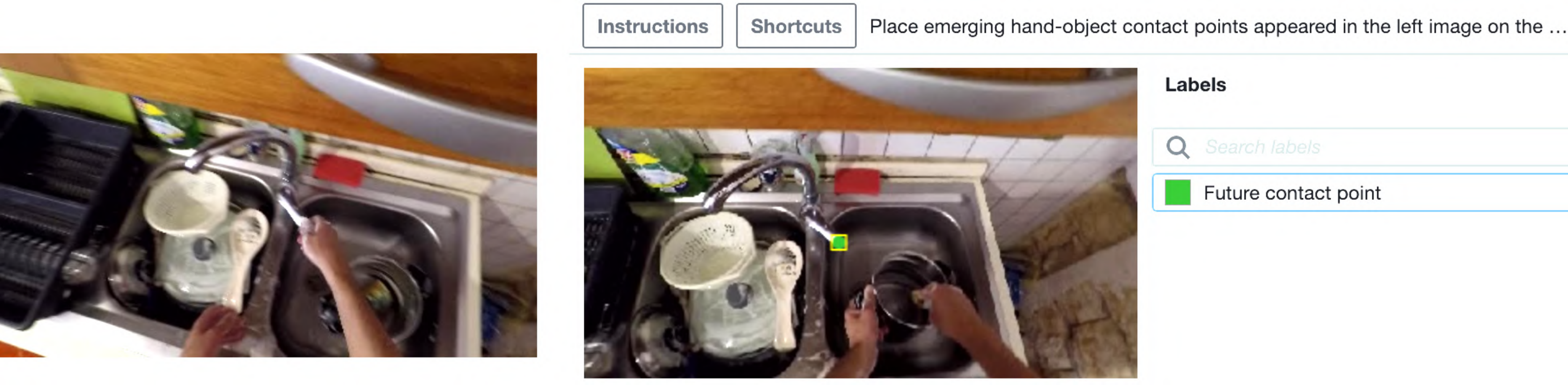}
\caption{Interface for collecting interaction hotspot annotations on evaluation set. The left image shows the contact frame, while the right image shows the last observation frame. users are asked to place points (\textcolor{tgreen}{green} dots) on the same object location in the right image touched by the hand in the left image. All the placed points are visible and on some objects.}
\label{fig: eval_amazon}
\end{figure}

We use the Amazon Mechanical Turk platform to collect interaction hotspot annotations on the evaluation set. The interface is shown in Figure~\ref{fig: eval_amazon}. We provide the contact frame in the left image and the last observation frame in the right image in the layout. Users are asked to place points on the same object location in the right image touched by the hand in the left image. The green dots are the labeled points by users. We require all the placed points to be visible and touched by the hand in the left image but haven't been touched in the right image. We collect 1-5 contact point labels for each sample in the evaluation set.

\begin{table}[t]
\scriptsize
\tablestyle{7pt}{1.2}
\begin{tabular}{ccc}\shline
 Stage & Configuration & Output\\\shline
 0 & Input videos & $T \times 256\times 454 \times 3$\\\shline
  & {\textbf{Backbone}}  & \\ \hline
 \multirow{1}*{1} & \multirow{1}*{TSN~\cite{wang2016temporal}} & $T \times 1024 \times 8 \times 14$\\ \hline
  \multirow{1}*{1} & \multirow{1}*{Hand-RoiAlign~\cite{he2017mask}} & $2 \times T \times 1024$\\ \hline
  \multirow{1}*{1} & \multirow{1}*{Object-RoiAlign~\cite{he2017mask}} & $2 \times T \times 1024$\\ \hline
  \multirow{1}*{1} & \multirow{1}*{Global-RoiAlign~\cite{he2017mask}} & $ T \times 1024$\\ \hline
  
 & {\textbf{Hand-Object Detector}~\cite{shan2020understanding}}  & \\ \hline
 \multirow{1}*{1} & \multirow{1}*{Hand location} & $2 \times T \times 4$\\ \hline
 \multirow{1}*{1} & \multirow{1}*{Object location} & $2 \times T \times 4$\\ \hline

 & {\textbf{Pre-processing}}  & \\ \hline
 \multirow{1}*{2} & Hand MLP & $2 \times T \times 512$ \\ \shline
 \multirow{1}*{2} & Object MLP & $2 \times T \times 512$ \\ \shline
 \multirow{1}*{2} & Global MLP & $T \times 512$ \\ \shline
 \multirow{1}*{2} & Input tokens & $5 \times T \times 512$ \\ \shline
 
 & {\textbf{OCT Encoder} $\mathcal{E}$}  & \\ \hline
 \multirow{1}*{3} & \multirow{1}*{encoding blocks $\mathcal{B}$} & $ 5 \times T\times 512$\\ \hline
 & {\textbf{OCT Decoder} $\mathcal{D}$}  & \\ \hline
\multirow{1}*{4} & \multirow{1}*{decoding blocks $\mathcal{B}$} & $5 \times T\times 512$\\ \hline
 & {\textbf{Hand C-VAE}} & \\ \hline
 \multirow{1}*{5} & \multirow{1}*{encoding function $\mathcal{F}_{enc}$} & $256$ ($\mu$)\\ 
 \multirow{1}*{5} & \multirow{1}*{decoding function $\mathcal{F}_{dec}$} & $2$ ($\mathcal{H}$)\\ \shline 
 & {\textbf{Object C-VAE}} & \\ \hline
 \multirow{1}*{5} & \multirow{1}*{encoding function $\mathcal{F}_{enc}$} & $256$ ($\mu$)\\ 
 \multirow{1}*{5} & \multirow{1}*{decoding function $\mathcal{F}_{dec}$} & $2$ ($\mathcal{O}$)\\
 \shline
 
\end{tabular}
\caption{Network architecture of the proposed model and corresponding dimensionality.}
\label{supp: network-arch}
\end{table}

\begin{table}[t]
\centering
\caption{\textbf{Comparison against 3D CNN} on EK100 dataset. ($\uparrow$/$\downarrow$ indicates higher/lower is better.) The 3D CNN uses I3D with ResNet-50 as backbone architecture. The proposed transformer model outperforms 3D CNN in both trajectory estimation and interaction hotspots prediction across all metrics.}
\label{tab: exp-3dcnn}
\footnotesize
\setlength{\tabcolsep}{5pt}
\begin{tabular}{c cc ccc}
\toprule
\multicolumn{1}{c}{} & \multicolumn{2}{c}{Trajectory} & \multicolumn{3}{c}{Interaction Hotspots}\\
\cmidrule(lr){2-3} \cmidrule(lr){4-6}
Model & ADE  $\downarrow$ & FDE  $\downarrow$ & SIM $\uparrow$ & AUC-J$\uparrow$ & NCC $\uparrow$ \\
\shline
I3D~\cite{carreira2017quo} & 0.19 & 0.16 & 0.16 & 0.64 & 0.55 \\
Ours & \textbf{0.12} & \textbf{0.11} & \textbf{0.19} & \textbf{0.69} & \textbf{0.72}\\
\bottomrule
\end{tabular}
\end{table}

\begin{table}[t]
\centering
\caption{\textbf{Comparison of performance by adopting end-to-end training} of our model on the EK100 dataset. In the paper, we report the performance of utilizing the frozen backbone. Here we compare the performance by training the model end-to-end. We observe a slight performance gain on trajectory estimation. Given that they achieve comparable performance and training end-to-end is more time-consuming, we freeze the backbone in our experiments.}
\label{tab: exp-end2end}
\footnotesize
\setlength{\tabcolsep}{5pt}
\begin{tabular}{c cc ccc}
\toprule
\multicolumn{1}{c}{} & \multicolumn{2}{c}{Trajectory} & \multicolumn{3}{c}{Interaction Hotspots}\\
\cmidrule(lr){2-3} \cmidrule(lr){4-6}
End-to-End & ADE  $\downarrow$ & FDE  $\downarrow$ & SIM $\uparrow$ & AUC-J$\uparrow$ & NCC $\uparrow$ \\
\shline
No & 0.12 & \textbf{0.11} & \textbf{0.19} & 0.69 & \textbf{0.72}\\
Yes & \textbf{0.11} & \textbf{0.11} & \textbf{0.19} & \textbf{0.70} & 0.70\\
\bottomrule
\end{tabular}
\end{table}

\section{Implementation Details}
\label{supp: implementation}
In our proposed Transformer model, we set the embedding dimension to be $512$ and use a dropout rate of $0.1$ for both encoding and decoding blocks. In the C-VAE head network of both hand and object, we implement it as a $2$-layer MLP, each for the encoding function $\mathcal{F}_{enc}$ and the decoding function $\mathcal{F}_{dec}$.  In the regular training epochs, we use cosine annealed learning rate decay starting from $1e-4$. During inference, as we need the hand location in the last observation frame as the $0$-th input to the decoder, we set the normalized left hand location to $(0.25, 1.5)$ and right hand location to $(0.75, 1.5)$ when any of them are invisible, followed~\cite{kapidis2019egocentric}. Our model is implemented with PyTorch~\cite{paszke2019pytorch}.

\paragraph{Epic-Kitchens.} On Epic-Kitchens, our model takes $2.5$s observations as input and forecasts future $1$s hand trajectory and interaction hotspots. We sample the videos at $4$ fps for training and evaluation. We train our model for $35$ epochs, including $5$ epochs warmup.

\paragraph{EGTEA Gaze+.} On EGTEA Gaze+, we set the anticipation time to be $0.5$s following~\cite{liu2020forecasting, girdhar2021anticipative}, given it has a smaller angle of view against the Epic-Kitchens dataset. Our model takes $1.5$s observations as input. We sample the videos at $6$ fps for training and evaluation. We train our model for $25$ epochs, including $5$ epochs warmup.

\section{Network Architectures}
\label{supp: network}

The network architecture is illustrated in Table~\ref{supp: network-arch}. We utilize ROIAlign~\cite{he2017mask} to crop the global, hand, and object features in each input frame $t$ with dimension $1024$. Then the extracted features and the detected hand and object bounding box locations are fused in the pre-processing module to get Transformer input tokens. The tokens are passed through the OCT encoder and decoder independently. The final future hand trajectory $\mathcal{H}$ at each time step is sampled from the hand C-VAE in an auto-regressive manner. The final object contact points are similarly sampled from the object C-VAE.

\section{Additional Experiments}
\label{supp: exp}

\paragraph{Comparison to 3D CNNs.} We compare our proposed Transformer model with 3D CNN, which is widely used in video understanding. We adopt the I3D~\cite{carreira2017quo} with ResNet-50~\cite{he2016deep} as backbone for 3D CNN. On the top of the backbone output, we predict the future hand locations and contact points by two head networks, similar to the hand and object head used in OCT. The I3D is pre-trained on Kinetics~\cite{kay2017kinetics} dataset. We trained the 3D CNN under the same setting as we trained OCT. The performance is shown in Table~\ref{tab: exp-3dcnn}. Experimental results show that by utilizing Transformer architecture against 3D CNN, we could improve the performance on both tasks. The OCT improves the FDE by $58.3\%$ and ADE by $45.5\%$ for trajectory estimation, SIM by $+3\%$, AUC-J by $+5\%$, and NSS by $+17\%$ for hotspots prediction on the EK100 dataset. This demonstrates the superiority of adopting Transformer architecture for visual forecasting.

\paragraph{End-to-end training.} In the main paper, we freeze the backbone TSN~\cite{wang2016temporal} and only train the OCT. We compare the performance against training end-to-end by fine-tuning the backbone along with the OCT. We apply data augmentation including random flipping and color jittering during training. The performance is shown in Table~\ref{tab: exp-end2end}. As can be seen, both models achieve comparable performance on both tasks. Given that training end-to-end is more time-consuming, we freeze the backbone in our experiments to accelerate training.

\begin{figure}[t]
\centering
\includegraphics[width=\linewidth]{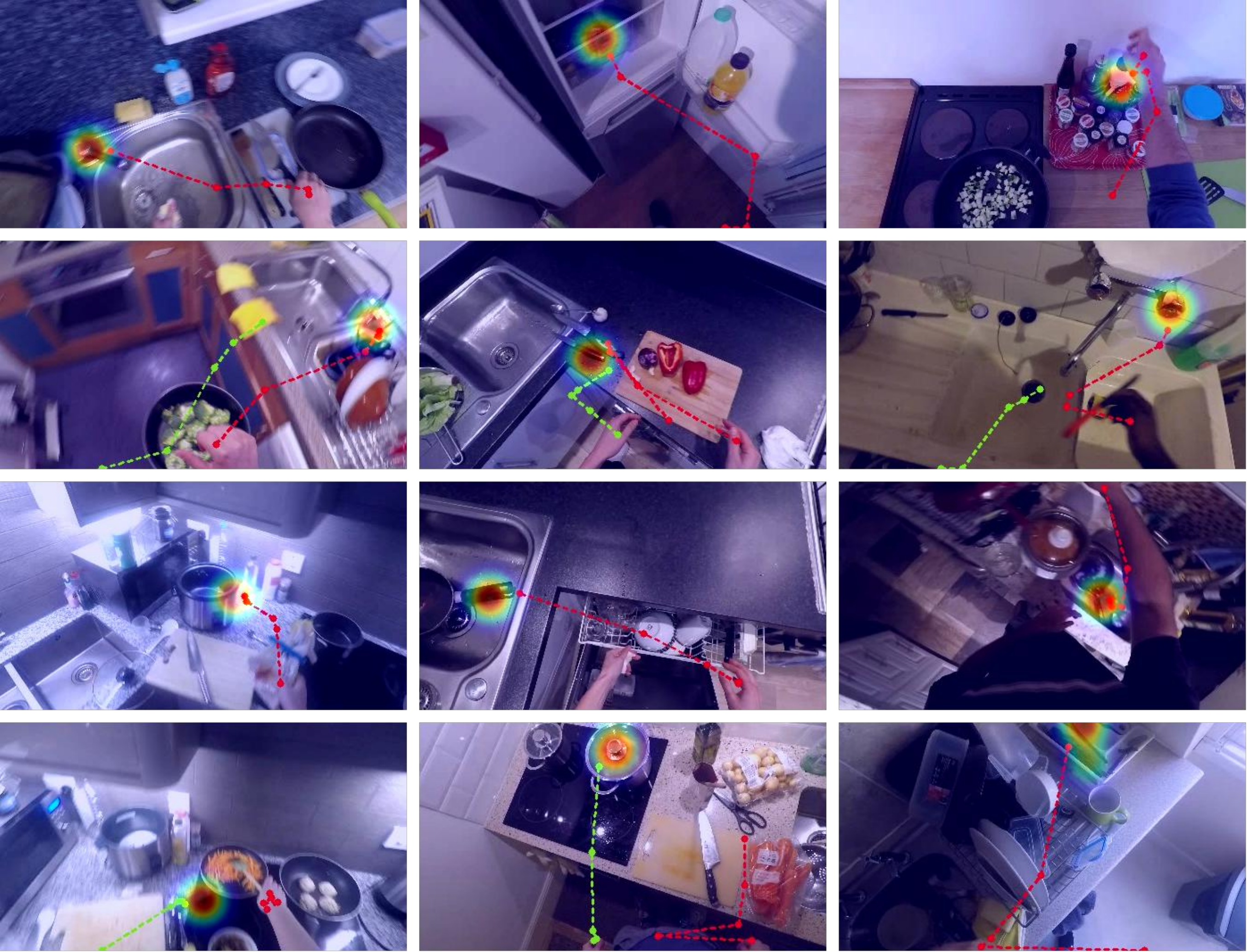}
\caption{Visualization of the automatically generated training labels on Epic-Kitchens dataset. The right and left future hand trajectory are shown in \textcolor{red}{red} and \textcolor{green}{green}. The heatmap indicates the interaction hotspots.}
\label{fig: vis_ek_labels}
\end{figure}

\begin{figure}[t]
\centering
\includegraphics[width=\linewidth]{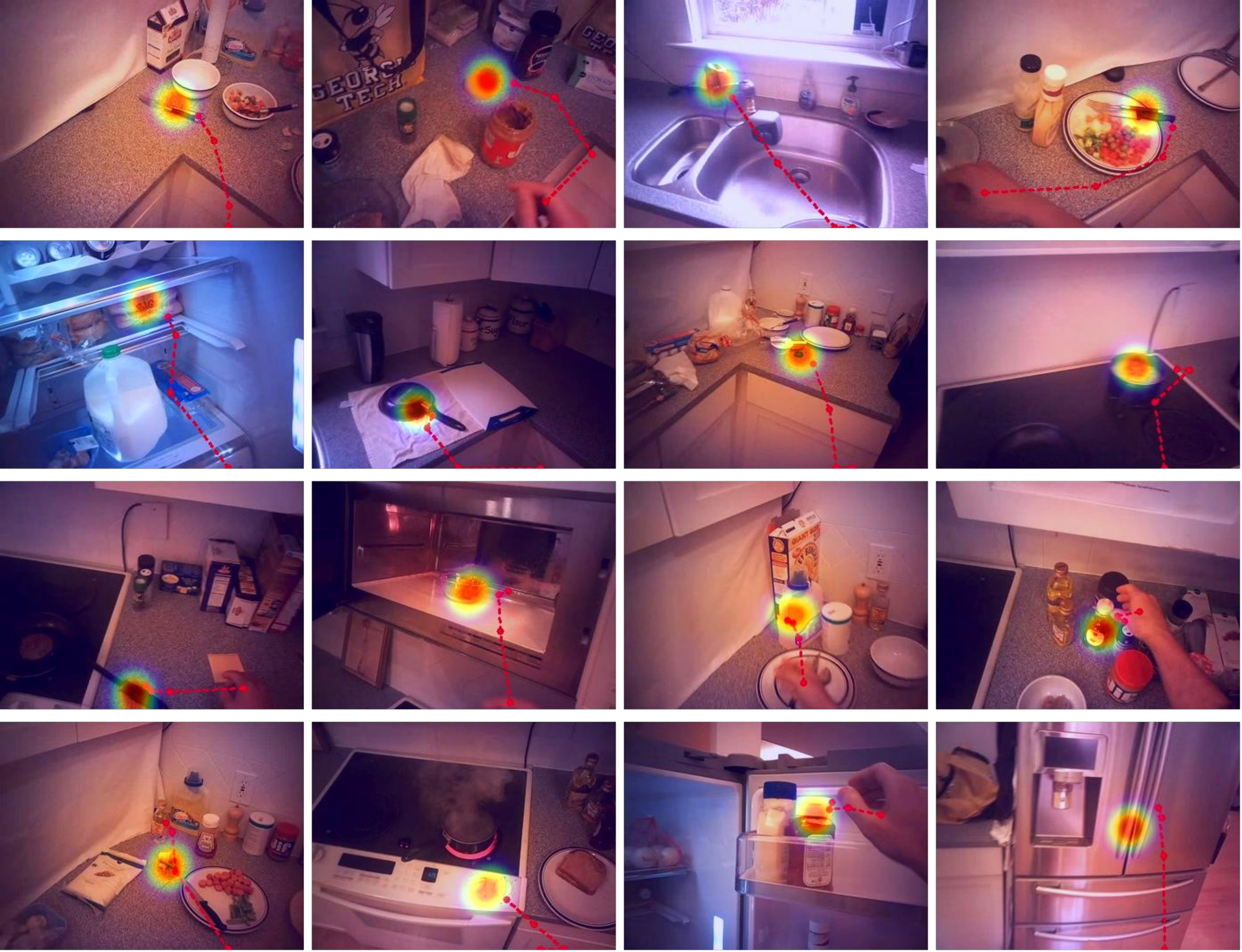}
\caption{Visualization of the automatically generated training labels on EGTEA Gaze+ dataset. The right and left future hand trajectory are shown in \textcolor{red}{red} and \textcolor{green}{green}. The heatmap indicates the interaction hotspots.}
\label{fig: vis_egtea_labels}
\end{figure}

\begin{figure}[t]
\centering
\includegraphics[width=\linewidth]{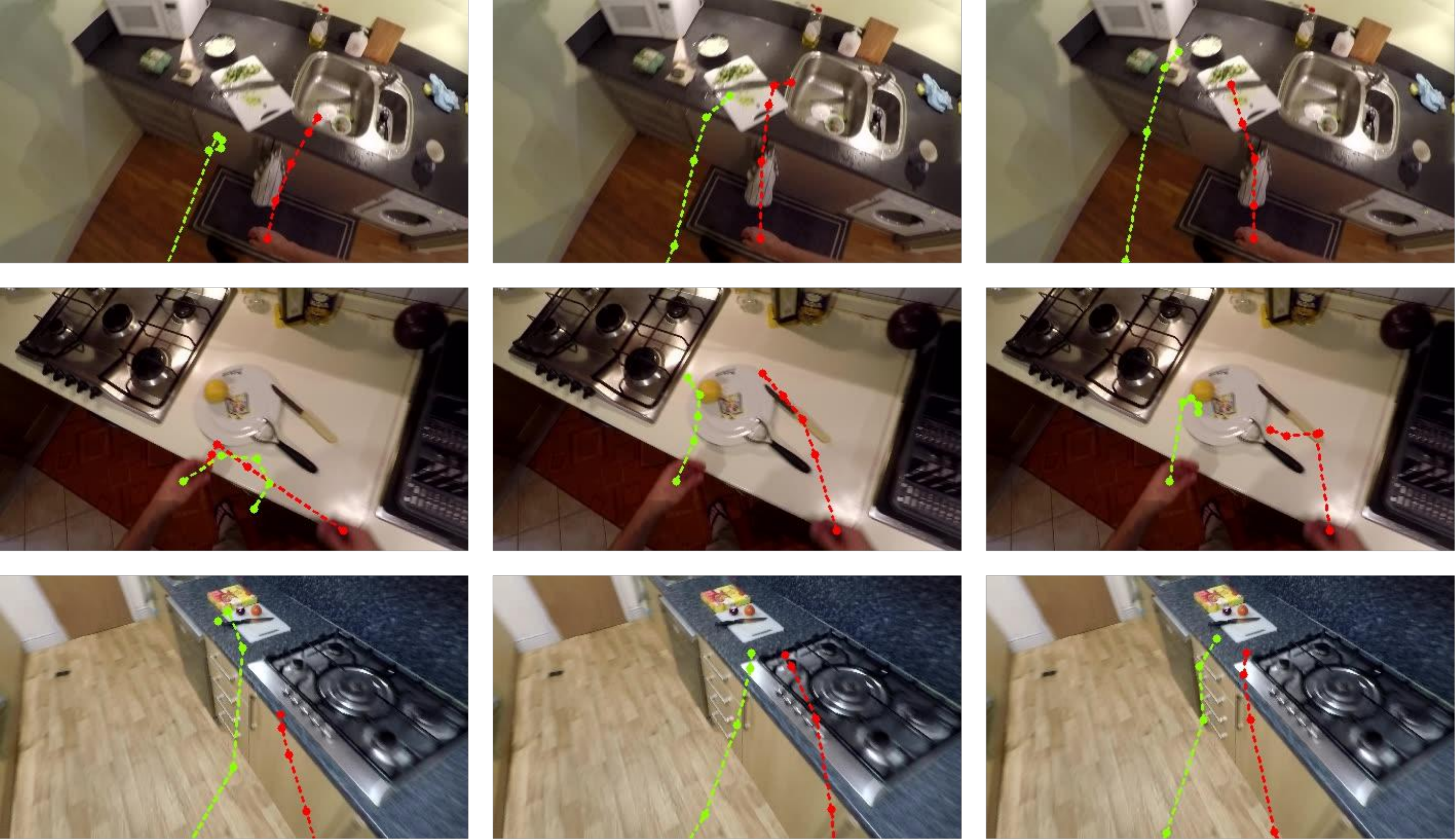}
\caption{Qualitative comparison of future hand trajectory against \textbf{Seq2Seq}~\cite{sutskever2014sequence} on the EK100 dataset. The \textbf{first column} shows the \textbf{Seq2Seq prediction}, the \textbf{second column} shows results of \textbf{our method}, and the \textbf{third column} are the \textbf{ground-truth}. The right and left hand trajectory are shown in \textcolor{red}{red} and \textcolor{green}{green}. Our method's prediction is more close to ground-truth against Seq2Seq and better reflects human's intention.}
\label{fig: vis_traj}
\end{figure}

\begin{figure}[t]
\centering
\includegraphics[width=\linewidth]{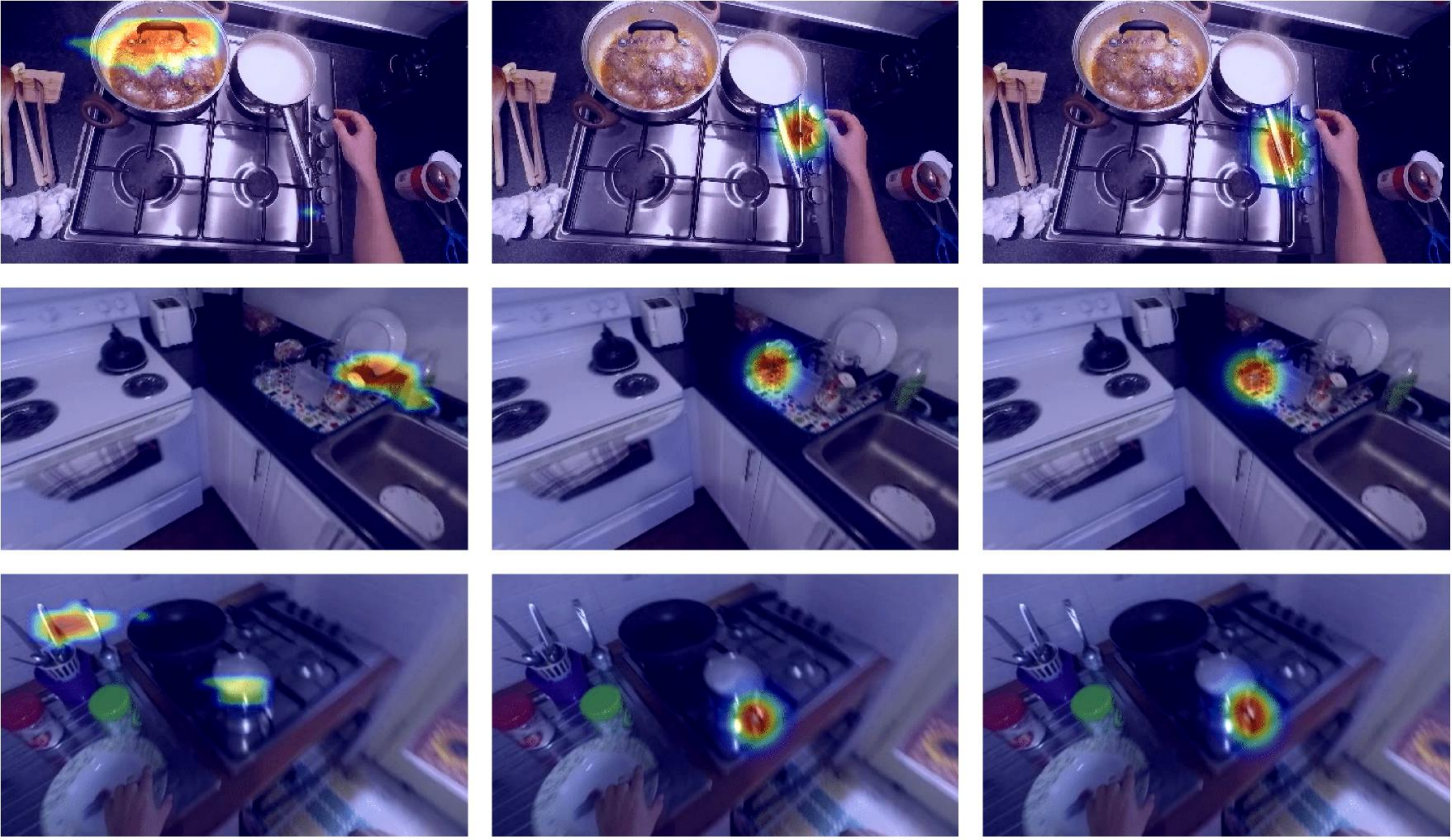}
\caption{Qualitative comparison of interaction hotspots estimation against \textbf{Hotspots}~\cite{nagarajan2019grounded} on EK100 dataset. The \textbf{first column} shows the \textbf{Hotspots prediction}, the \textbf{second column} shows results of \textbf{our method}, and the \textbf{third column} are the \textbf{ground-truth}. From the
visualization, we observe Hotspots fails when there are multiple candidate objects in the cluttered scene, while our method could better capture the future interactions.}
\label{fig: vis_hotspot}
\end{figure}

\begin{figure}[t]
\centering
\includegraphics[width=\linewidth]{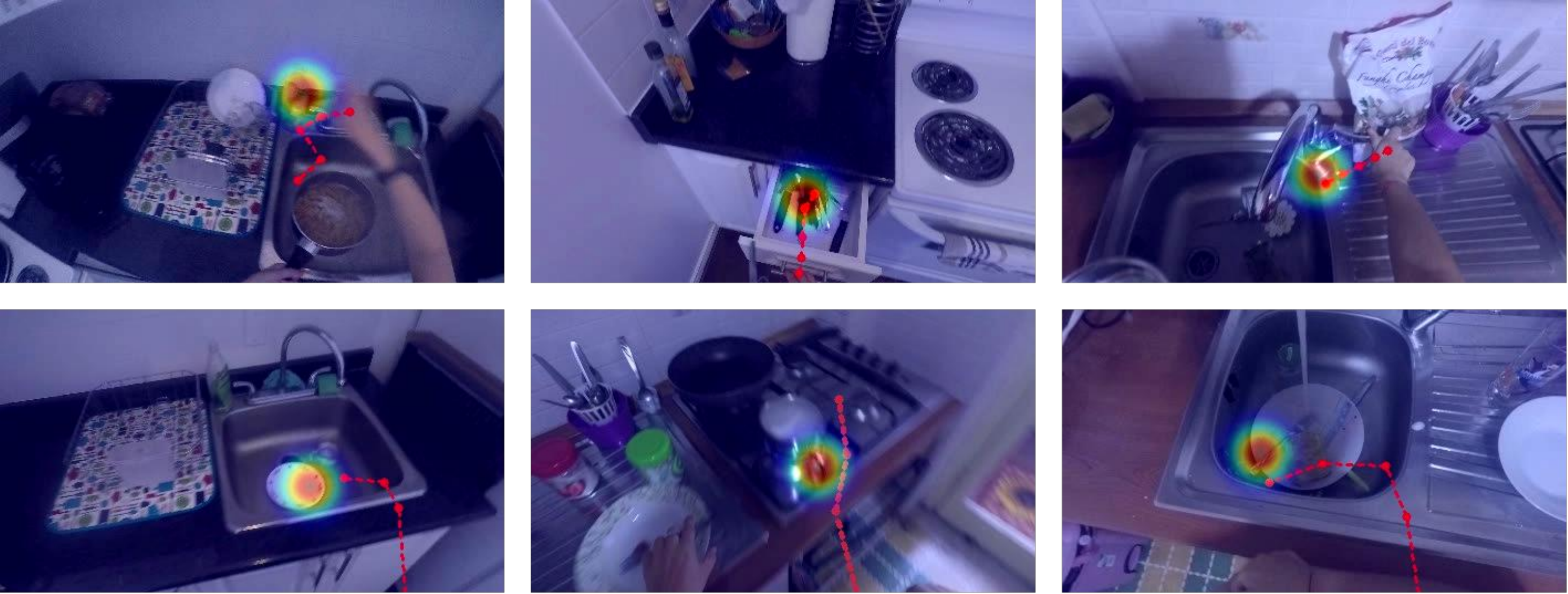}
\caption{Qualitative visualization of future hand trajectory and interaction hotspots on unseen kitchens and participants on the EK100 dataset. Our model is generalizable to unseen environments and could give reasonable predictions.}
\label{fig: vis_unseen}
\end{figure}

\begin{figure}[t]
\centering
\includegraphics[width=\linewidth]{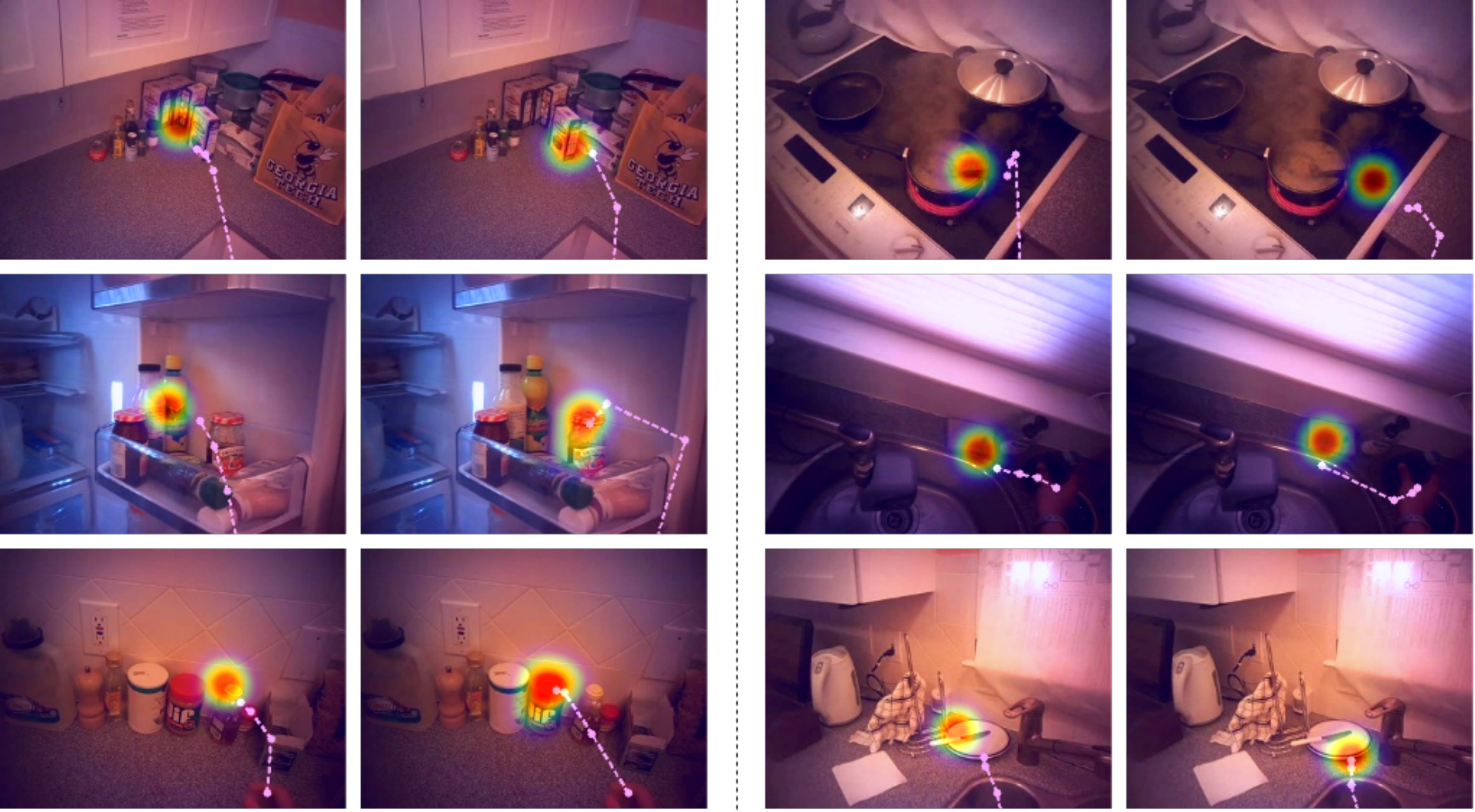}
\caption{Qualitative visualization of cross-dataset generalization results. The model is trained on Epic-Kitchens and tested on EGTEA Gaze+. We show 6 different samples. In each pair of sample, the \textbf{left} shows our model prediction, the \textbf{right} shows the ground-truth. The future hand trajectory is shown \textcolor{purple}{purple}. Our model demonstrates strong cross-domain generalization.}
\label{fig: vis_cross_gen}
\end{figure}

\section{Training Labels Visualization}
\label{supp: vis-train}
We visualize the automatically generated training labels on Epic-Kitchens and EGTEA Gaze+ datasets in Figure~\ref{fig: vis_ek_labels} and Figure~\ref{fig: vis_egtea_labels}. It can be seen from the figures that our method could generate high-quality training labels under different kitchen environments and different subjects.

\section{Qualitative Comparisons}
\label{supp: vis} 
We compare our model's prediction of future hand trajectory and interaction hotspots against methods that achieved second-best performance in each task, as reported in Table~\ref{tab: exp-traj} and Table~\ref{tab: exp-obj} in the main paper. 

\paragraph{Hand trajectory comparison.} We visualize the prediction results on the EK100 of our method and SeqSeq~\cite{sutskever2014sequence} that utilizes LSTM for trajectory estimation. The results are shown in Figure~\ref{fig: vis_traj}. As can be seen, our method's prediction is more close to the ground-truth against Seq2Seq and better reflects human intention. 

\paragraph{Object interaction hotspots comparison.} We compare our model's prediction of interaction hotspots against Hotspots~\cite{nagarajan2019grounded} that employ  Grad-Cam~\cite{selvaraju2017grad} to infer future hotspots map. Note that Hotspots takes the ground-truth future action label and last observation frame as input. The results are shown in Figure~\ref{fig: vis_hotspot}. We observe that Hotspots's prediction is struggling when there are multiple objects present in a cluttered scene. This implies forecasting future interaction hotspots is more challenging than the video affordance grounding task solved by Hotspots. The future hotspots estimation needs observation frames as context to locate future hand-object interactions.

\section{Generalization Results Visualization}
\label{supp: vis-gen}

\paragraph{Generalization on the unseen kitchens.} We visualize our model's prediction on the unseen environment on the EK100 dataset. The selected samples come from the validation split that contains unseen kitchens and participants. We show $6$ different samples in Figure~\ref{fig: vis_unseen}. Though the kitchen environment is unseen in training, our model could still predict reasonable future hand trajectory and interaction hotspots, which shows the in-domain generalization ability of our model.

\paragraph{Cross-dataset generalization.} 
We visualize the cross-dataset generalization ability on the EGTEA Gaze+ dataset. The model is trained on Epic-Kitchens and tested on EGTEA Gaze+. We show $6$ different samples in Figure~\ref{fig: vis_cross_gen}. Our model could well capture the human intention under unseen environments and subjects, forecasting future hand trajectory and interaction hotspots close to the ground-truth. It demonstrates the strong cross-domain generalization ability of our model.

\end{document}